\documentclass[letterpaper, 10 pt, conference]{ieeeconf}  

\IEEEoverridecommandlockouts                              

\usepackage{amsmath,amssymb,amsfonts}
\usepackage{algorithmic}
\usepackage{graphicx}
\usepackage{textcomp}
\usepackage{booktabs}
\usepackage{subcaption}

\usepackage{tabularx}
\usepackage{svg}
\usepackage{url}
\usepackage{xcolor}
\newcommand{\new}[1]{\textcolor{black}{#1}}
\newcommand{\zvec}{\mathbf{z}}
\newcommand{\xvec}{\boldsymbol{\varphi}}
\newcommand{\R}{\mathbb R}
\newcommand{\uvec}{\mathbf u}
\newcommand{\vvec}[1]{\Bigr[\def\arraystretch{0.8}\begin{array}{cc}#1\end{array}\Bigr]}
\newcommand{\bd}[1]{\mathcal B(#1)}
\newcommand{\Su}{\mathbf S_{\mathbf u}}
\newcommand{\inn}[1]{\left\langle #1\right\rangle}

\newcommand{\cU}{\mathcal U}

\DeclareMathOperator*{\argmin}{arg\,min}
\newcommand{\hmin}{h_{\textrm{min}}}
\newcommand{\cH}{\mathcal H}

\newcommand{\norm}[1]{\left\lVert{#1}\right\rVert}
\newcommand{\ntrain}{n_{\textrm{train}}}
\newcommand{\mnys}{m_\textrm{nys}}

\usepackage{booktabs}
\newcommand{\tableheadline}[1]{\textbf{#1}}

\title{\LARGE\bf Dynamic Robotic Cloth Folding with Efficient Koopman Operator-Based Model Predictive Control
}

\author{Edoardo Caldarelli\textsuperscript{1}, Franco Coltraro\textsuperscript{2}, Adrià Colomé\textsuperscript{2}, Lorenzo Rosasco\textsuperscript{1, 3}, and Carme Torras\textsuperscript{2}%
\thanks{\textsuperscript{1}Istituto Italiano di Tecnologia, Genoa, Italy \textsuperscript{2}Institut de Robòtica i Informàtica Industrial, CSIC--UPC, Barcelona, Spain \textsuperscript{3}MaLGa Center, DIBRIS, Università degli Studi di Genova, Genoa, Italy. Correspondence to: \texttt{edoardo.caldarelli@iit.it}.}
\thanks{This work was partially funded by the  European Project HORIZON-CL4-2024-DIGITAL-EMERGING-01-101189600 (FlexCycle). E.\ Caldarelli acknowledges support from COST Action InterCoML (CA24136). F.\ Coltraro was fully supported by Momentum CSIC
Programme project MMT24-IRII-01 and now is partially by ClothIRI (CSIC 202350E080) project. L.\ Rosasco acknowledges the financial support of the European Commission (Horizon Europe grant ELIAS 101120237), and the Ministry of Education, University and Research (FARE grant ML4IP R205T7J2KP).}
}
\begin{document}

\maketitle
\thispagestyle{empty}
\pagestyle{empty}

\begin{abstract}

Robotic cloth folding is a challenging task, particularly {when considering dynamic folding tasks, which aim at folding cloth by fast motions that leverage its dynamics.} 
{When subject to such fast motions,} the complexity of cloth dynamics hinders both system identification and planning of folding trajectories, resulting in a difficult simulation-to-reality transfer when using physical models of cloth. Compared to the dexterity that humans exhibit when performing folding tasks, robotic approaches usually employ small garments with quite rigid dynamics, and are either too slow, or fast but imprecise, requiring several attempts to achieve a reasonably good fold. In this paper, we tackle 
{these challenges by generating fast folding trajectories with a novel model predictive controller, integrating physics-based simulation of cloth dynamics and efficient, kernel-based Koopman operator regression.}
Koopman operator regression, an increasingly popular machine learning technique for nonlinear system identification, is used to obtain a linear model for the cloth being folded. Such a surrogate model, trained with data from a high-fidelity, physics-based cloth simulator, can then be employed within a suitable model predictive control algorithm, in place of the costly, nonlinear one, to efficiently generate folding trajectories to be executed by a robotic manipulator. Both in simulated and real-robot experiments, we show how the linearization supplied by the Koopman operator-based model can be employed to efficiently generate fast folding trajectories to unseen poses, without sacrificing folding accuracy.
\end{abstract}

\section{Introduction {and Related Works}}
\begin{figure}[t]
\centering
\includegraphics[width=\linewidth]{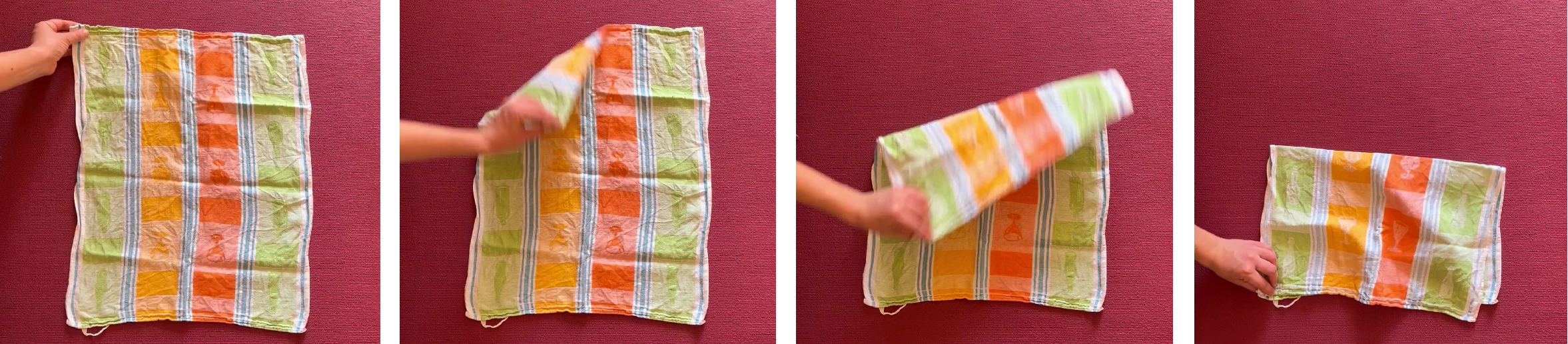}
\caption{An example of a dynamic cloth folding trajectory performed by a human expert. Albeit efficient, the high speed of the motion may hinder the accuracy of the fold, if the cloth dynamics are not taken into consideration.}
\label{fig:real_folding}
\end{figure}

Robotic cloth manipulation is an increasingly relevant area of research, which greatly challenges classic robot control algorithms \cite{yin2021modeling,longhini2024unfolding}. The deformable nature of cloth requires state-of-the-art techniques in order to simulate \cite{coltraro2022inextensible,coltraro2024novel} and control \cite{ha2022flingbot,avigal2022speedfolding,hietala2022learning,longhini2024adafold} the cloth dynamics within a robotic platform, with the ultimate goal of autonomously performing cloth manipulation skills with the same degree of expertise humans exhibit in their everyday lives, {as shown, e.g., in Fig.\ \ref{fig:real_folding}}.\looseness=-1

Within the context of robotic cloth manipulation, we can distinguish between \emph{quasi-static} tasks, and \emph{dynamic} manipulation skills \cite{mason1993dynamic}. In this paper, we focus on the latter case: We propose an algorithm for \emph{cloth folding} that leverages the dynamics of the cloth, to completely fold a rectangular cloth by acting only on one of its corners. Namely, we aim at performing such a task with a single manipulator and a single robot motion \cite{hietala2022learning,longhini2024adafold}, therefore considering a more resource-efficient environment compared to bimanual setups, or single-handed, quasi-static ones, which typically require multiple motion primitives \cite{blanco2023qdp}. {Besides domestic and assistive environments, dynamic manipulation is of particular importance in industrial settings where robots could automatize the manual folding of cloth. Speed is critical for efficiency as evidenced by the fact that trained humans that work in textile factories always fold clothes dynamically and in a very fast fashion.}

In our setting, dynamics-aware skills become of paramount importance, due to the high inertia of the cloth when manipulated fast. {These inertial phenomena are known to cause} severe sim-to-real gaps \cite{blanco2024benchmarking}. Nonetheless, as shown in our experiments, we are able to achieve a successful {zero}-shot sim-to-real transfer on fast folding actions (less than 1.5 s), by deploying the {control strategy} described below.
\begin{figure*}
    \centering    
    \includegraphics[width=\linewidth]{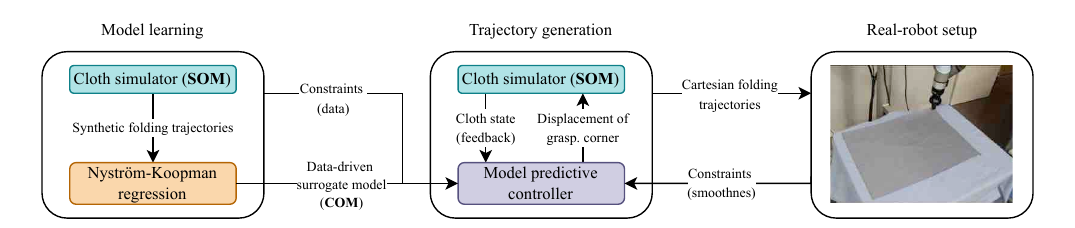}
	\caption{Summary of the cloth folding {strategy} proposed in this paper. A cloth simulator \cite{coltraro2024novel} simulates synthetic folding trajectories, which are used as training data for a Koopman operator regression model \cite{caldarelli2025linear}. The data-driven model, along with suitable constraints, are fed to a model predictive controller. In this way, a folding trajectory to a given target pose is generated, accounting for performance requirements (speed and accuracy of the fold) as well as constraints stemming from the real robot setup (e.g., smoothness of the velocities). The trajectory is then executed on the real robot.  
	}
	\label{fig:summary}
\end{figure*}
The dynamic folding skills are achieved {by means of a novel model predictive control (MPC) framework, which integrates both an accurate, and a data-driven efficient model of the cloth dynamics. Specifically, high-fidelity folding data are generated with a physics-based simulator \cite{coltraro2024novel}, and use to train a non-parametric machine learning approximation, obtained through \emph{Koopman operator regression} \cite{koopman1931hamiltonian,korda2018linear,bevanda2021koopman,caldarelli2025linear}}. Koopman operator regression recasts the nonlinear dynamics of cloth to linear ones (albeit in an infinite dimensional vector space), that can be learned efficiently \cite{caldarelli2025linear} and embedded in a \textit{linear} MPC pipeline. MPC can then be used to generate trajectories that allow the robot to fold the cloth to unseen target poses. Most importantly, those trajectories account for \emph{performance criteria} and \emph{user-defined} constraints (as opposed to the training data), which can be easily embedded in the optimal control problem underlying MPC.

{In our control architecture}, the high fidelity simulator from \cite{coltraro2024novel} {acts} as a \emph{simulation-oriented model} (SOM), whereas the Koopman approximation from \cite{caldarelli2025linear} serves as a \emph{control-oriented model} (COM). The SOM can be used to provide accurate feedback in simulation, whereas the COM can be used to perform cheap, but reliable forecasts of the cloth dynamics without needing any control or position derivatives of the SOM. Solving the MPC with these two models yields a folding trajectory that can then be executed with a real robot, to accomplish the folding task. {The integration of the components yielding our cloth folding strategy is detailed} in Fig.\ \ref{fig:summary}.\looseness=-1

\smallskip

\paragraph*{Contributions} In summary, in this paper we:
\begin{itemize}
    \item combine the physics-based, collision-aware SOM from \cite{coltraro2024novel} with the efficient Koopman operator regression algorithm studied in \cite{caldarelli2025linear}, obtaining a data-driven, linear COM for a piece of cloth;
    \item use such a COM in a linear MPC control strategy, to generate constrained robot trajectories, in {closed-loop with the SOM}, folding a piece of cloth to an unseen target pose; 
    \item execute and evaluate the generated trajectories on a real robotic platform, showcasing how \new{our methodology reduces the sim-to-real gap}. {Most importantly, the folding motions happen in less than 1.5 s, counteracting the inertial effects of cloth due to such a high speed.}
\end{itemize}

\paragraph*{Outline} The structure of this paper is as follows: Section \ref{sec:case_study} describes the setup we consider in this work, reporting the details of the SOM for the cloth. Section \ref{sec:controller_design} details the COM, and the data-driven MPC controller we propose in this paper. Section \ref{sec:experiments} assesses the proposed control pipeline in simulated and real robot experiments. Lastly, Section \ref{sec:conclusions} contains the conclusions.
\section{Physical Model of the Cloth}
\label{sec:case_study}
In this section we give a self-contained presentation of the inextensible cloth model developed in \cite{coltraro2022inextensible,coltraro2024novel} so that it can be used to generate the data needed to train the Koopman operator regressor, {which will be in turn used as the backbone of a novel MPC strategy for dynamic cloth folding}. 

We assume that the piece of cloth $S$ we wish to control has been discretized into a quadrilateral mesh and the position of all its $N$ vertices or nodes (denoted by $p_i(t) = [x_i(t),y_i(t),z_i(t)]^\intercal\in\mathbb{R}^3$) at time $t\geq 0$ is given by $\boldsymbol{\varphi}(t) = (\textbf{x}(t),\textbf{y}(t),\textbf{z}(t))^\intercal\in\mathbb{R}^{3N}$, where $\textbf{x}(t),\textbf{y}(t),\textbf{z}(t)$ denote the $x$-coordinates (resp. $y$ and $z$) of all the nodes of the  discrete surface in $\mathbb{R}^{N}$ at time $t$ and $\textbf{v}^\intercal$ denotes the tranpose of $\textbf{v}$. 

\medskip

Then, the inextensible cloth model including aerodynamics but without collisions consists in the following set of ordinary differential equations (ODEs):
\begin{equation}\label{ODE_aero}
	\begin{cases}
		\rho\textbf{M}\ddot{ \boldsymbol{\varphi}} =  -\delta\textbf{M}\textbf{g}  - \kappa\textbf{K} \boldsymbol{\varphi} - \alpha\textbf{M}\dot{ \boldsymbol{\varphi}} - \nabla\textbf{C}( \boldsymbol{\varphi})^\intercal \boldsymbol{\lambda}\\
		\textbf{C}( \boldsymbol{\varphi}) = 0,
	\end{cases}
\end{equation}
where the meaning of each term and parameter (see also Table \ref{tabla:0}) is:
\begin{itemize}
\item $\rho>0$ is the density of the cloth (assumed to be homogeneous, see Table \ref{table:sota}) and $\textbf{M}$ is the (diagonal) mass matrix where each $m_k > 0$ is one fourth of the sum of the areas of all incident quads to the node $p_k$,


\item the term $-\delta\textbf{M}\textbf{g}$  accounts for the force of gravity, where $\textbf{g} = [0,\dots,0|0,\dots,0|g,\dots,g]^\intercal$  and $g = 9.8\text{m}/\text{s}^2$. In order to model the aerodynamic effects of air resistance on cloth we allow the inertial $\rho$ and gravitational $\delta$ masses to be different (for a detailed discussion and justification of this simplified aerodynamics model, see \cite{coltraro2025aero}) introducing thus an upwards constant lift force. Hence, we set $\delta \leq \rho$ as a new parameter of the model which we call the \textit{virtual mass},


\item the stiffness matrix (we are using the isometric bending model described in \cite{Bergou:2006:QBM}) is $\textbf{K} = \textbf{L}^\intercal\textbf{M}\textbf{L}$
where $\textbf{L}$ is an approximation of the point-wise Laplacian and $\kappa>0$ is a bending constant,


\item the term $\alpha\textbf{M}\dot{ \boldsymbol{\varphi}}$ is called Rayleigh damping and the magnitude of $\alpha > 0$ is responsible for modeling the dampening of slow oscillations of the cloth and the drag of air \cite{Zienkiewicz:2005:FEM},

    
\item and finally $ \boldsymbol{\lambda}(t)$ are the Lagrange multipliers ensuring \textit{inextensibility} (to be described next) and other possible positional constraints, in our case manipulating the textile by prescribing the position of one of its corners.
\end{itemize}
 
Besides the aforementioned positional constraints introduced by manipulation, the smooth function $\textbf{C}: \mathbb{R}^{3N}\rightarrow\mathbb{R}^{n_C}$ is responsible for modeling the \textit{inextensibility} of the textile through the satisfaction of the equality constraints $\textbf{C}(\boldsymbol{\varphi}) = \textbf{0}$. Each constraint $C_i(\boldsymbol{\varphi}(t))=0,\;i=1,\dots,n_C$ is in fact a quadratic function of its argument and we have $n_C$ of them, depending on the number of nodes $N$ of the discretization (for more details, see \cite{coltraro2022inextensible}).
\begin{table}[t]
	\centering
	\begin{tabularx}{0.4\textwidth}{Xl} \toprule
		\tableheadline{Parameter}       & \tableheadline{Meaning}\\ \midrule
		$\rho$                          & Density (inertial mass) \\
		$\delta$                        & Virtual (gravitational)  mass   \\
		$\kappa$                        & Bending/stiffness            \\
		$\alpha$                        & Damping of slow oscillations \\
		\bottomrule
	\end{tabularx}
	\caption{Physical parameters of the inextensible cloth model \new{relevant for the folding task considered} and their meaning. 	\label{tabla:0}}
\end{table}

The inextensibility constraints $\textbf{C}(\boldsymbol{\varphi}) = \textbf{0}$ model what is usually called the \textit{internal dynamics} of cloth; furthermore, for its application in our scenario, we also need to include collisions of the cloth with the table and with itself. We model this by enforcing a set $H_i(\boldsymbol{\varphi}(t))\geq 0$, $i = 1,\ldots, n_H$ of $n_H$ inequality constraints $\textbf{H}( \boldsymbol{\varphi}) \geq \textbf{0}$, such that {the differentiable function $\textbf{H}: \mathbb{R}^{3N}\rightarrow\mathbb{R}^{n_H}$ satisfies that for each $i=1,\dots,n_H$} we have a {non-zero} outwards normal $\nabla H_i( \boldsymbol{\varphi})\neq \textbf{0}$ (e.g. for a table locally $\nabla H_i(p_k) = (0,0,1)$). We can then model (self)collisions together with friction by including non-smooth  forces into the equations of motion of the cloth (this is known as Signorini's contact model, see \cite{coltraro2024novel} for more details). 

\subsection{Numerical Integration of the System} \label{sec:num}
To integrate the system numerically from time $t_j$ to $t_{j+1} = t_j + \Delta t$ given positions $\boldsymbol{\varphi}_{j}$ and velocities $\dot{\boldsymbol{\varphi}}_{j}$ at $t_j$, we perform an iterative process  
\begin{equation}
    \boldsymbol{\varphi}^{i+1} =  \boldsymbol{\varphi}^i + \Delta \boldsymbol{\varphi}^{i+1}\label{eq:discrete_dynamics}
\end{equation}
where the initial point is the unconstrained step $ \boldsymbol{\varphi}^0$ given by applying an implicit Euler scheme to the equations of motion (\ref{ODE_aero}) ignoring all equality and inequality constraints. Moreover, we write:
\begin{equation*}
\textbf{H}( \boldsymbol{\varphi}^{i+1})= \textbf{H}( \boldsymbol{\varphi}^i + \Delta \boldsymbol{\varphi}^{i+1})\simeq \textbf{H}( \boldsymbol{\varphi}^{i}) + \nabla\textbf{H}( \boldsymbol{\varphi}^{i})\Delta \boldsymbol{\varphi}^{i+1},
\end{equation*}
and similarly 
\begin{equation*}
\textbf{C}( \boldsymbol{\varphi}^{i+1})= \textbf{C}( \boldsymbol{\varphi}^i + \Delta \boldsymbol{\varphi}^{i+1})\simeq \textbf{C}( \boldsymbol{\varphi}^{i}) + \nabla\textbf{C}( \boldsymbol{\varphi}^{i})\Delta \boldsymbol{\varphi}^{i+1},
\end{equation*}
and then solve iteratively for $\textbf{q} := \Delta \boldsymbol{\varphi}^{i+1}$ the following sequence of quadratic programs with linear equality and inequality constraints:
\begin{equation}\label{prob_qua}
\begin{cases}
	\min_{\textbf{q}}\tfrac{1}{2}\textbf{q}^\intercal\cdot\textbf{M}\cdot \textbf{q} - \textbf{q}^\intercal\cdot\textbf{f}_{\mu}(\dot{ \boldsymbol{\varphi}}^i)  \\
	\textbf{C}( \boldsymbol{\varphi}^{i}) + \nabla\textbf{C}( \boldsymbol{\varphi}^{i})\textbf{q} = \textbf{0},\\
	\textbf{H}( \boldsymbol{\varphi}^{i}) + \nabla\textbf{H}( \boldsymbol{\varphi}^{i})\textbf{q} \geq \textbf{0},
\end{cases}
\end{equation}
where $\dot{\boldsymbol{\varphi}}^{i}=\frac{\boldsymbol{\varphi}^{i}-\boldsymbol{\varphi}_{j}}{\Delta t}$ is the approximation of $\dot{\boldsymbol{\varphi}}_{j+1}$ at iteration $i$ and $\textbf{f}_{\mu}(\dot{ \boldsymbol{\varphi}}^i)$ accounts for Coloumb's friction. For more theoretical and implementation details, see \cite{coltraro2024novel}. 
\subsection{Parameter Estimation for the Cloth and a Priori Formulas}
The previously described physical model shows great accuracy in describing a wide variety of real cloth motions including very dynamic movements and fast collisions within a margin of error of 1 cm (see \cite{coltraro2022inextensible} for an experimental validation with depth cameras and \cite{coltraro2024novel} for one with a motion capture system). However, by virtue of introducing the aerodynamics parameter $\delta > 0$ (the virtual mass), the optimal value of the physical parameters of Table \ref{tabla:0} varies not only with each different textile but also with the speed at which it moves (except for the density $\rho$ which is constant for each fabric). 

Therefore, in \cite{coltraro2025aero} \textit{a priori} formulas for the values of the parameters $\alpha$ and $\delta$ where developed:
\begin{align}\label{formula_aeroR1}
	\begin{split}
	\hat{\delta} &= -0.0223 -0.0178S + 0.0714 V + 0.7664\rho, \\ \hat{\alpha} &= +0.2082 -0.1481S + 1.1804 V + 1.7440\rho,
	\end{split}
\end{align}
where $\rho$ is the density of the cloth, $S$ is a normalized area measure (in our case it will be always $2$) and $V$ is the average (in time and over all nodes of the cloth) of $50\%$ of the highest squared velocities. Notice that this formula allows us to deduce the value of the cloth's parameters depending only on its density, size and speed. The coefficients of the linear formulas (\ref{formula_aeroR1}) were found by using optimization with several real \new{Motion Capture (MoCap)} recordings of textiles of different sizes moving at various speeds. In \cite{coltraro2025aero} it was shown that using formulas (\ref{formula_aeroR1}) gives a comparable absolute error to using the optimal values of $\alpha$ and $\delta$ found by minimizing cloth errors for each individual recording. 

Lastly, it is relevant to mention that in the \new{MoCap} recordings used for obtaining formulas (\ref{formula_aeroR1}), the stiffness $\kappa$ of the cloths had almost no influence in the motions and hence could not be estimated reliably. \new{Nevertheless, for the dynamic folding task considered in this paper, this parameter (along with $\alpha$ and $\delta$) has a big influence on the motion of the cloths. Although this parameter $\kappa$ could be estimated by using a suitable version of a simulated Cusick test, see \cite{coltraro2022inextensible}}, in this work we {empirically tune that parameter, based on the associated performance of the controller.}

\section{Proposed Controller Design}
\label{sec:controller_design}
The controller {deployed} in this article is an MPC approximation of the infinite-horizon LQR studied in \cite{caldarelli2025linear}. 
\subsection{Surrogate System Identification}
In this work, we rely on the system identification techniques developed in  
\cite{caldarelli2025linear}. There, the nonlinear data, obtained through the system's dynamics (in our case obtained by iteratively solving (\ref{prob_qua})), are lifted to an infinite-dimensional space, where the dynamics are linear through the \emph{Koopman operator}. Then, this infinite-dimensional space is approximated with a finite-dimensional one, through a data-based approach, i.e., the Nyström method. Specifically, we solve here a regularized least-squares problem, to learn a surrogate dynamical model for the cloth being folded. In this subsection, we give an overview of the steps taken to build such surrogate model, to later use it as a COM in a MPC in Section \ref{subsec:mpc}.\looseness=-1

\subsubsection{Learning problem} When considering the dynamics of the cloth being folded, we classically take as control input the displacement (i.e., the variation in position) of the corner grasped by the robot \cite{coltraro2022inextensible}. However, this modeling choice may be unrealistic for real-robot scenarios, since it requires to have a perfect point grasp of the corner considered. In this work, we relax this assumption by modeling a \emph{vertical grasp} of one corner, with fixed orientation of the robot's end-effector, i.e., we control the position of \emph{two} adjacent points in the cloth mesh. As we will show in this section, imposing such a constant orientation can be naturally encoded as a constraint for the MPC. In the following, the controlled displacement will be denoted as $\Delta \uvec\in\R^6$. The discretization described in Section \eqref{sec:num} allows to measure triples of the form $(\xvec_j, \Delta\uvec_j, \xvec_{j+1})$ where $\xvec_j$ represents the state of the vertices of the cloth mesh at time $t_j$. For a suitable flow map $f:\R^{3N}\times \R^{6}\to \R^{3N}$, we aim at building a data-driven surrogate for the transitions
\begin{equation}\label{eq:duscrete_time_ODE_sys}
    \xvec_{j+1} = f(\xvec_j, \Delta \uvec_j),
\end{equation}
for $j\geq 0$. This is done by means of Koopman operator regression with approximated reproducing kernels \cite{caldarelli2025linear}. 
\subsubsection{State transformation and regression problem} Let us consider a reproducing kernel Hilbert space $\cH$, associated to the kernel function $k:\R^{3N}\times \R^{3N}\to\R$. Let $\psi:\R^{3N}\to\cH$ be the corresponding canonical feature map. After having collected a set of $\ntrain$ training inputs (i.e.,\ a tuple $(\xvec_i, \Delta\uvec_i)$) and training outputs (i.e., $\xvec_{i}^+ = f(\xvec_i, \Delta\uvec_i)$), we set up the following regression problem, to be solved over the set of linear operators from $\cH\times \R^6$ to $\cH$:
\begin{equation}
    G = \arg\min_{\mathbf W:\cH\times \R^6\to\cH}\mathcal R(\mathbf W) + \gamma \norm{\mathbf W}^2_{\textrm{HS}},\label{eq:erm}
\end{equation}
where 
\begin{equation}
    \mathcal R(\mathbf W) = \frac1n\sum_{i=1}^{\ntrain}\norm{\psi(\xvec_i^+) - \mathbf W\vvec{\psi(\xvec_i) \\\Delta\uvec_i}}^2_{\cH}.
\end{equation}
As shown by \cite{caldarelli2025linear}, \eqref{eq:erm} can be approximated with the Nyström method, to improve the computational efficiency of the solution \cite{nystrom1930praktische,williams2000using}.  
\subsubsection{Nyström approximation} The Nyström approach entails sub-sampling $\mnys$ points from the training set, named \emph{Nyström landmarks} and denoted as $\{\tilde \xvec_i\}_{i=\{1\dots \mnys\}}$. Let $\boldsymbol \Pi:\cH\to\cH$ be the orthogonal projector on $\mathrm{span}\{k(\xvec_1, \cdot), \dots, k(\xvec_{\mnys}, \cdot)\}$, and let $\boldsymbol \Pi_\textrm{in}:\cH\times\R^6\to \cH\times\R^6$ be the block diagonal projector $\boldsymbol \Pi_\textrm{in}\vvec{\psi(\xvec)\\\Delta \uvec} = \vvec{\boldsymbol\Pi\psi(\xvec)\\\Delta \uvec}$. Problem \eqref{eq:erm} is replaced by
\begin{equation}
    \mathbf G = \arg\min_{\mathbf W:\cH\times \R^6\to\cH}\mathcal R(\boldsymbol\Pi \mathbf W\boldsymbol\Pi_\textrm{in}) + \gamma \norm{\mathbf W}^2_{\textrm{HS}}.\label{eq:nys}
\end{equation}
Solving this problem allows to retrieve a linear operator that evolves the dynamics of the transformed state forward in time. Namely, for an initial state $\xvec_0$,
\begin{equation}
    \begin{cases}
         z_0 = \psi(\xvec_0),\\
         z_{t + 1} = \mathbf G\vvec{z_t\\\Delta\uvec_t}.\label{eq:linear_surrogate_dynamics}
    \end{cases}
\end{equation}
\subsubsection{Surrogate vector-valued dynamics} As shown in \cite{caldarelli2025linear}, the dynamics in \eqref{eq:linear_surrogate_dynamics} can be equivalently expressed in terms of vectors rather than functions in $\cH$. The closed-form expressions are given as follows. Let $\bd{\cdot, \cdot}$ be a block-diagonal operator s.t.\ $\bd{\mathbf M, \mathbf N} = \vvec{\mathbf M &0\\ 0&\mathbf N}$. We can further define $\delta_{i,j} = 1 \iff i=j$, $\alpha >0$, and the following kernel matrices:
\begin{itemize}
    \item $\mathbf k_{m, \xvec} \in \R^{\mnys}, (\mathbf k_{m, \xvec})_i = k(\tilde\xvec_i, \xvec)$;
    \item $\mathbf K_{m} \in \R^{\mnys\times \mnys}, (\mathbf K_{m, m})_{i, j} = k(\tilde\xvec_i, \tilde\xvec_j) + \alpha\delta_{i,j}$;
    \item $\Su \in \R^{\ntrain \times 3}$, $(\Su)_{i} = \frac{1}{\sqrt n}\uvec_i^\intercal$;
    \item $\mathbf K_{nm} \in \R^{\ntrain\times \mnys}$, $(\mathbf K_{nm})_{i,j}=k(\xvec_i, \tilde\xvec_j) $
    \item $\mathbf K_{mn} = \mathbf K_{nm}^\intercal$.
\end{itemize}
For initial conditions
\begin{equation}
    \zvec_0 = (\mathbf K_{m}^{\dagger})^{1/2}\mathbf k_{m, \xvec_0},\label{eq:initial_state}
\end{equation}
we can define the following data-driven system for $t \geq 0$ \cite{caldarelli2025linear}:
\begin{align}
    \zvec_{t+1} &=  (\mathbf K_{m}^{\dagger})^{1/2}\mathbf K_{mn}\vvec{\mathbf K_{nm} &\sqrt n \mathbf S_{\uvec}}\nonumber\\
    &\quad\cdot\left({\vvec{\mathbf K_{mn}\\ \sqrt n\Su^T}
				\vvec{ \mathbf K_{nm} &\sqrt n \Su }
				+ \gamma n \bd{
					\mathbf K_{m} , I } }\right)^{\dagger}\nonumber\\
                    &\quad\cdot\bd{\mathbf K_{m}^{1/2}, I}\vvec{\zvec_t\\ \uvec_t}.\label{eq:koopman_dynamics}
\end{align}
\subsubsection{State reconstruction} After defining these data-driven dynamics, we may be interested in reconstructing an approximated value for the state $\xvec$ from the surrogate state $\zvec$. Following \cite{caldarelli2025linear}, we do so by means of the following reconstruction matrix, obtained after stacking the regression outputs in the matrix $\mathbf X^+\in\R^{3N\times n}$:
\begin{equation}
    \mathfrak C = \mathbf X^+\mathbf K_{nm}
	(\mathbf K_{mn}\mathbf K_{nm} + \lambda n \mathbf K_{m})^{-1}\mathbf K_{m}^{1/2}.\label{eq:rec_mat}
\end{equation}
\subsection{Model Predictive Controller}
\label{subsec:mpc}
The dynamics introduced in the previous section are linear in the surrogate state $\zvec$ and in the control variable, and can be leveraged to define an optimal control problem (OCP), to be solved within an LMPC loop \cite{korda2018linear}. {In this section, we detail our novel MPC strategy}. Let $\mathfrak C$ be as defined in \eqref{eq:rec_mat}. Furthermore, let $\mathbf Q = \mathfrak C^\intercal\mathbf Q'\mathfrak C$, $\mathbf Q'\in\mathbb R^{3N\times 3N}$, $\mathbf R\in\R^{6\times 6}$, and $\xvec_r$ a target pose of the cloth. Lastly, let $\cU$ be a convex set containing the admissible control values. Then, we can define the following finite-horizon OCP, to be solved at time-step $\kappa$:
\begin{align}
    &\Delta\uvec^*_0, \dots,\Delta\uvec^*_{T}=\nonumber \\
    &\argmin_{\Delta\uvec_0, \dots, \Delta \uvec_{ T}}\sum_{t=0}^{T}\inn{(\zvec_t-\zvec_r), \mathbf Q(\zvec_t-\zvec_r)} 
    + \inn{\Delta\uvec_t, \mathbf R\Delta\uvec_t}\nonumber\\
    &\textit{subject to:}\nonumber\\
    &\zvec_0 = (\mathbf K_{m}^{\dagger})^{1/2}\mathbf k_{m, \xvec_\kappa},\nonumber\\
    &\zvec_r = (\mathbf K_{m}^{\dagger})^{1/2}\mathbf k_{m, \xvec_r},\nonumber\\
                       &\zvec_{t+1} =  (\mathbf K_{m}^{\dagger})^{1/2}\mathbf K_{mn}\vvec{\mathbf K_{nm} &\sqrt n S_{\uvec}}\nonumber\\
    &\qquad\cdot\left({\vvec{\mathbf K_{mn}\\ \sqrt n\Su^T}
				\vvec{ \mathbf K_{nm} &\sqrt n \Su }
				+ \gamma n \bd{
					\mathbf K_{m} , I } }\right)^{\dagger}\nonumber\\
                    &\qquad\cdot\bd{\mathbf K_{m}^{1/2}, I}\vvec{\zvec_t\\ \uvec_t},\nonumber\\
                       &\Delta\uvec_t \in \cU,\ \forall t \in \{0, \dots, T\}.\label{eq:ocp}
\end{align}
The position at time-step $\kappa$ to which the grasped points of the cloth need to be moved is obtained as
\begin{equation}
    \uvec_\kappa = \uvec_{\kappa-1} + \Delta \uvec_0^*.
\end{equation}
The control $\uvec_\kappa$ is applied to the SOM. Then, problem \eqref{eq:ocp} is solved again, with $\xvec_{\kappa + 1}$ as initial state, obtained from the SOM. 
\subsubsection{Constraints} 
The OCP in \eqref{eq:ocp} follows two types of constraints: Those related to the lifting of the state and dynamics, and those constraining the control actions, as described in the following.
\paragraph{Dynamics} The first three constraints appearing in \eqref{eq:ocp} correspond to the lifting of the cloth's state at the beginning of the OCP time window, the lifting of the reference state (i.e., the target mesh pose), and the dynamics in the lifted space, obtained through the data-driven approximation of the Koopman operator.
\paragraph{Constraining the control} One of the key advantages of MPC is the possibility of naturally embedding constraints in the OCP. This is of upmost importance considering that we are using a data-driven model, i.e., some regions of the control space (and consequently, of the state space) may not have been explored during training, and for trajectory shaping, as we will discuss in the following. Given the definition of the cloth dynamics proposed in Section \ref{sec:case_study}, we {we propose to} shape the control set $\mathcal U$ by defining the following constraints:
\begin{itemize}
    \item The OCP should not involve regions of the control space that were not sampled during the system identification phase, and for which the data-driven system becomes unreliable. Specifically, these constraints are of the form
    \begin{align}
        \uvec_{\kappa-1} + \Delta \uvec_{0} \geq& [-\infty, -y_{\textrm{min}}, \hmin]^T\\
        \uvec_{\kappa-1} + \Delta \uvec_{0}+\dots + \Delta \uvec_{t} \geq& [-\infty, -y_{\textrm{min}}, \hmin]^T,\nonumber\\
        &\forall t \in \{1, \dots T\},
    \end{align}
    to prevent the controlled corner to get too close to (or below) the table the cloth is placed on, along the vertical axis, as well as to fold towards the center of the cloth in the beginning. $y_{\textrm{min}}$ is the initial position of the grasped corner along the $y$ axis, $\hmin$ is the lowest initial height of the controlled points.
    \item Smoothness of the generated control trajectories should be enforced. This is of upmost importance considering that the trajectories of the grasped corner of the cloth, generated in simulation, need to be executed by a robotic manipulator, to perform the folding. For $s > 0$, these constraints are of the form 
    \begin{align}
    - s \leq &\Delta\uvec_0 - (\uvec_{\kappa -1}- \uvec_{\kappa-2}) \leq s,\\
    - s \leq &\Delta\uvec_t - \Delta\uvec_{t-1} \leq s,\ \forall t \in \{1, \dots, T\}.
    \end{align}
    \item The position of the two controlled points in the mesh should vary by the same displacement, to emulate the end-effector of the cloth moving with constant orientation. Let $\Delta\uvec_t^{[i:j]}$ be the $i$-to-$j$ entries of $\Delta\uvec_t$. This constraint can be encoded as
    \begin{equation}
        \Delta\uvec_t^{[1:3]} - \Delta\uvec_t^{[4:6]} = \mathbf 0
    \end{equation}
    \item Lastly, we should prevent the grasped corner from being displaced too far from its current position, avoiding unstable behaviors. For $\mathbf w\in \R^{6}$ and $\mathbf v\in \R^{6}$, these constraints are of the form
    \begin{equation}
        \mathbf w \leq \Delta\uvec_t \leq \mathbf v,\  \forall t \in \{0, \dots T\}.
    \end{equation}
\end{itemize}
\section{Experiments}
\label{sec:experiments}
\begin{figure}[t]
\centering
\includegraphics[width=.85\linewidth]{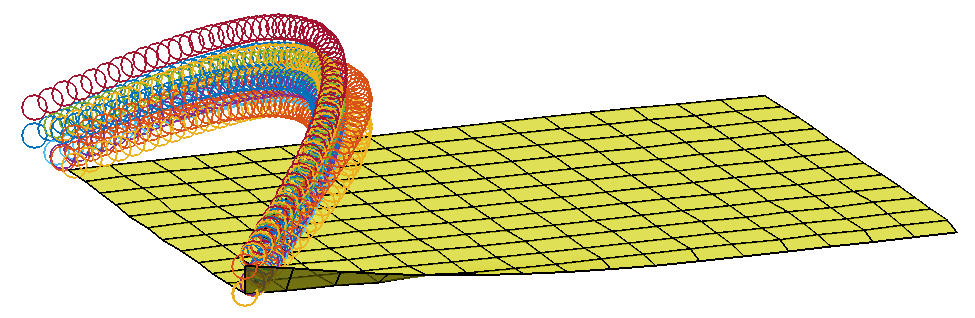}
\caption{Some of the training parabolas used for the data generation.}
\label{fig:training_parabolas}
\end{figure}
In this section, we evaluate the proposed control pipeline empirically, both in simulation and on a real robotic platform.
\begin{figure*}
\begin{subfigure}{.49\linewidth}
\centering
\includegraphics[width=\linewidth]{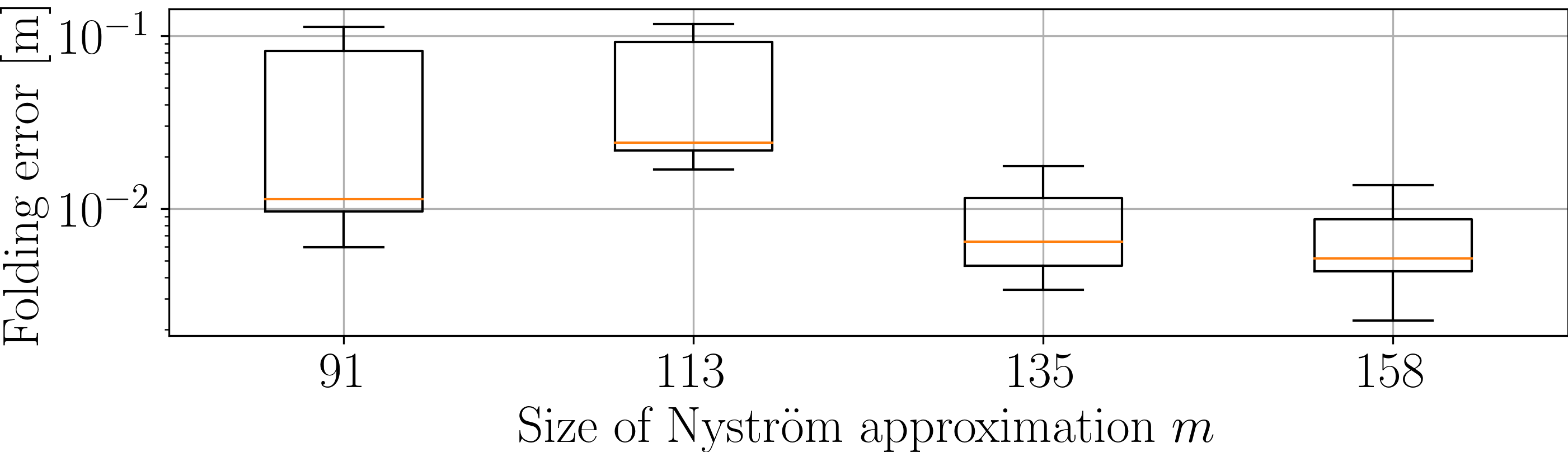}
\caption{Final folding error.}
\label{subfig:boxplots_mesh_error}
\end{subfigure}\hfill
\begin{subfigure}{.49\linewidth}
\centering
\includegraphics[width=\linewidth]{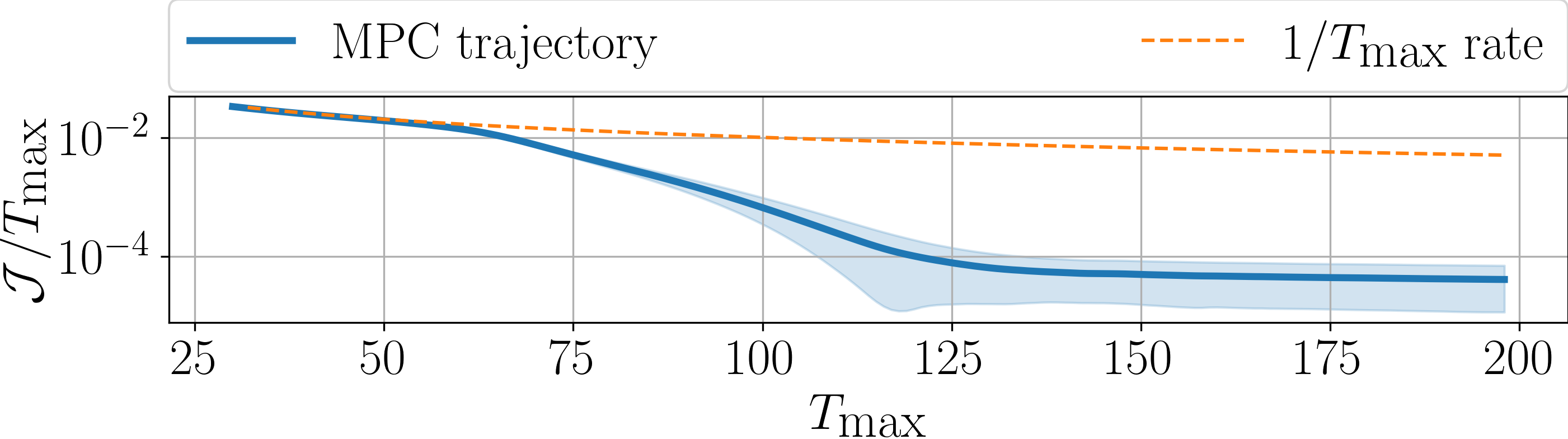}
\caption{Sublinear cost index $\mathcal J$.}
\label{subfig:boxplots_running_cost}
\end{subfigure}
\caption{(a) Final folding error evaluated across 10 target folds generated with single-handed motions, for different values of $m$. Increasing $m$ yields a more accurate COM, which in turn improves the performance of the MPC algorithm. (b) The running cost \eqref{eq:performance_index} (normalized in the interval $[0, 1]$), grows at a sublinear rate w.r.t.\ the horizon $T_{\textrm{max}}$. The MPC curve is obtained with $m= 158$ Nytröm landmarks. Mean $\pm$ std.\ across 10 target folds.}
\label{fig:state_size_effect}
\end{figure*}
\begin{figure*}[t]
\begin{subfigure}{.49\linewidth}
\centering
\includegraphics[width=\linewidth]{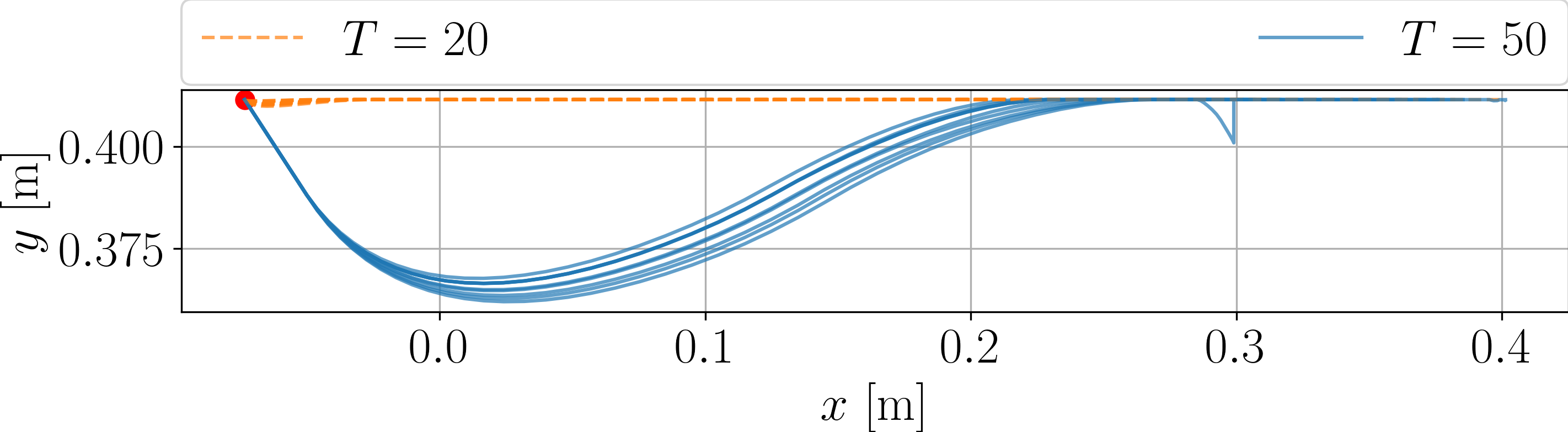}
\caption{Control trajectories for horizons $T$.}
\label{subfig:control_trajs_pred_hor}
\end{subfigure}\hfill
\begin{subfigure}{.49\linewidth}
\centering
\includegraphics[width=\linewidth]{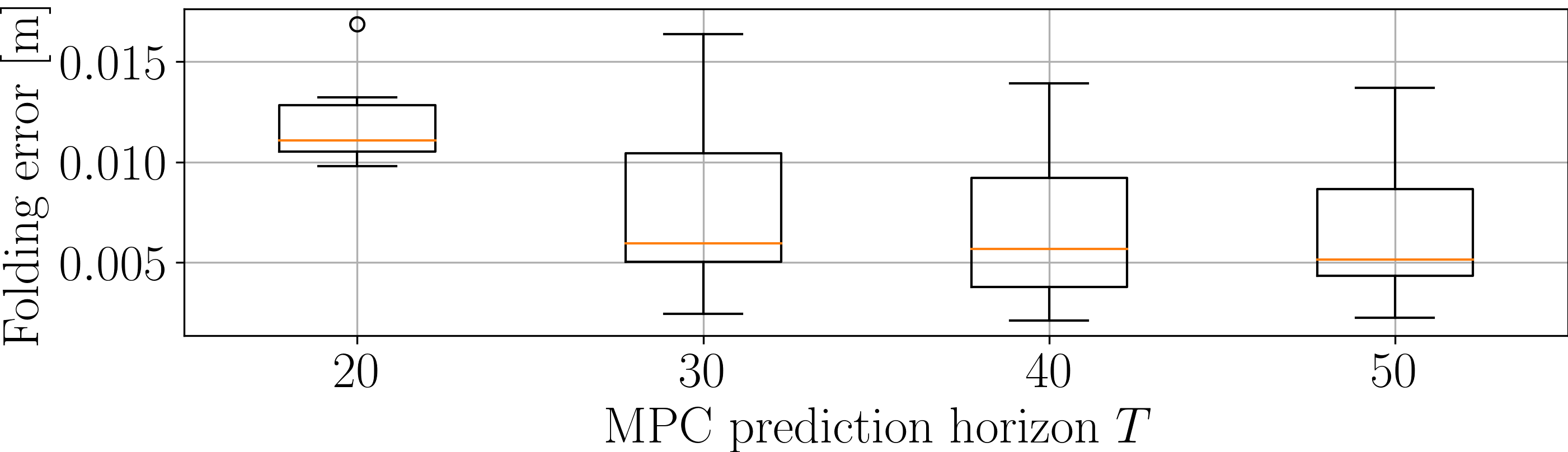}
\caption{Final mesh error $\mathcal M$.}
\label{subfig:boxplots_mesh_error_pred_hor}
\end{subfigure}
\caption{(a) $x$-$y$ visualization of the trajectories generated with different values of the prediction horizon $T$, and $m=158$. An initial motion towards the center of the cloth emerges as a consequence of a larger value for $T$. This type of motion is pivotal for the fold to be successful. The initial position is marked by a red dot. (b) Final folding error evaluated across 10 target folds generated with single-handed motions, for different values of the prediction horizon $T$. Increasing $T$ indeed yields a more effective control trajectory.}
\end{figure*}
 \begin{figure*}
\begin{subfigure}{.249\linewidth}
\centering
\includegraphics[width=\linewidth]{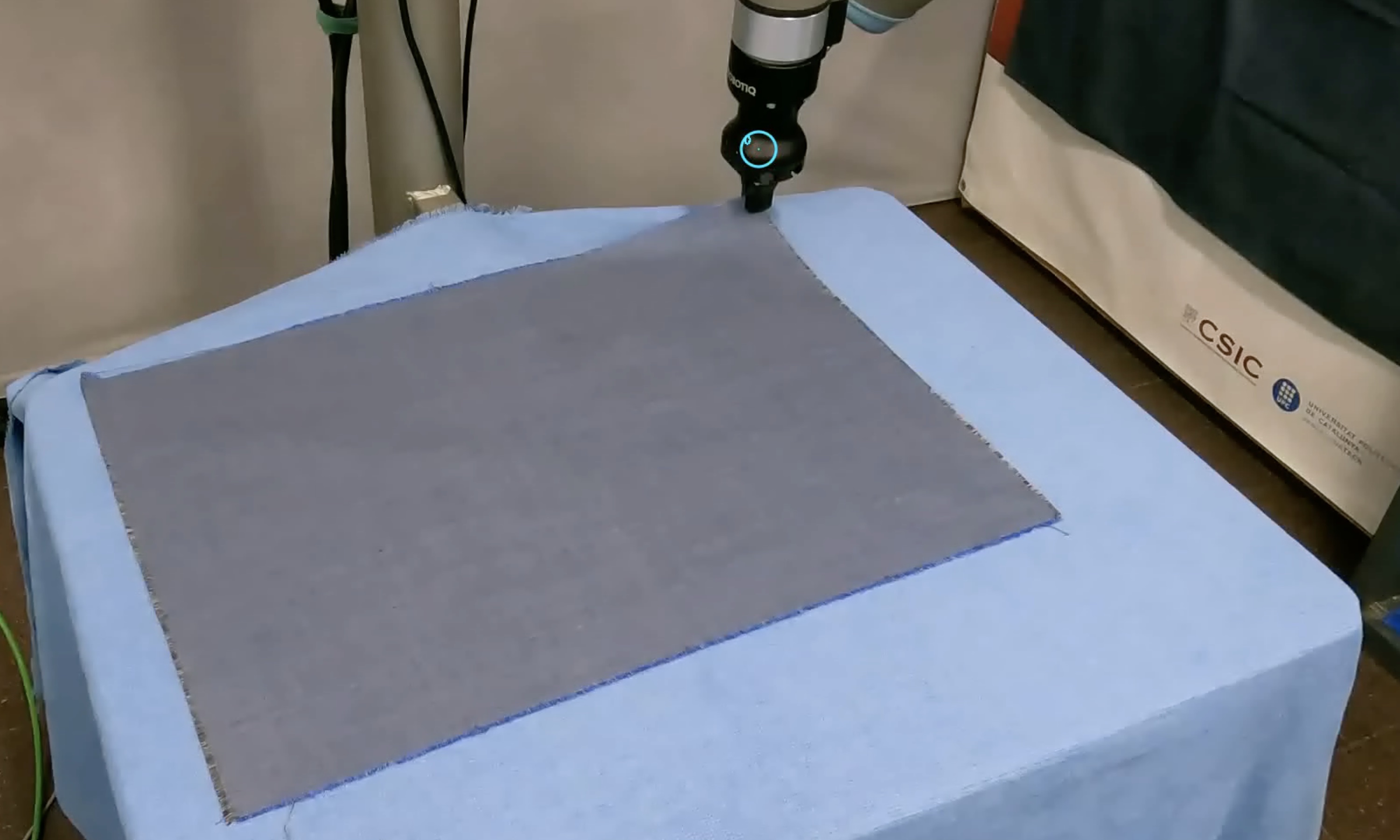}
\end{subfigure}\hfill
\begin{subfigure}{.249\linewidth}
\centering
\includegraphics[width=\linewidth]{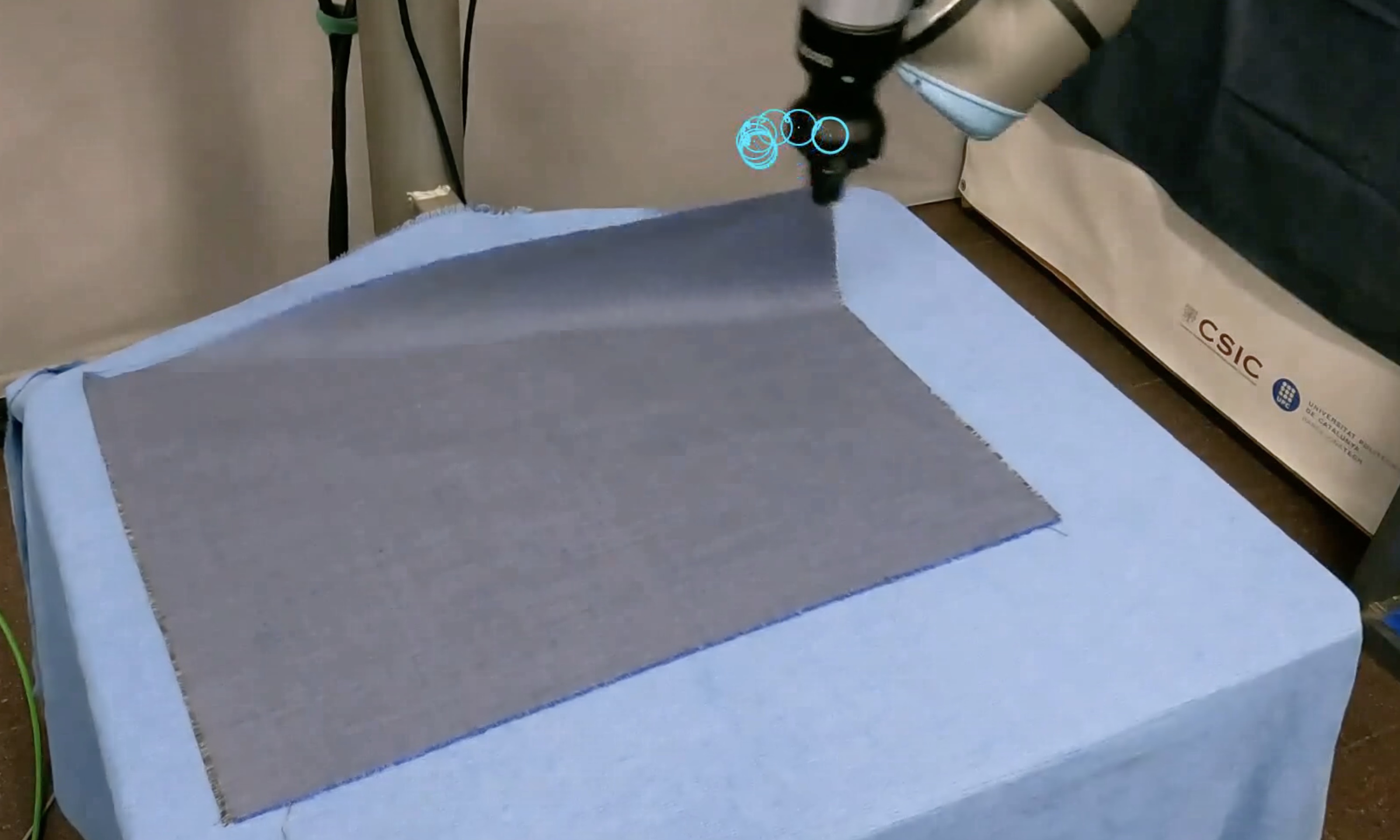}
\end{subfigure}\hfill
\begin{subfigure}{.249\linewidth}
\centering
\includegraphics[width=\linewidth]{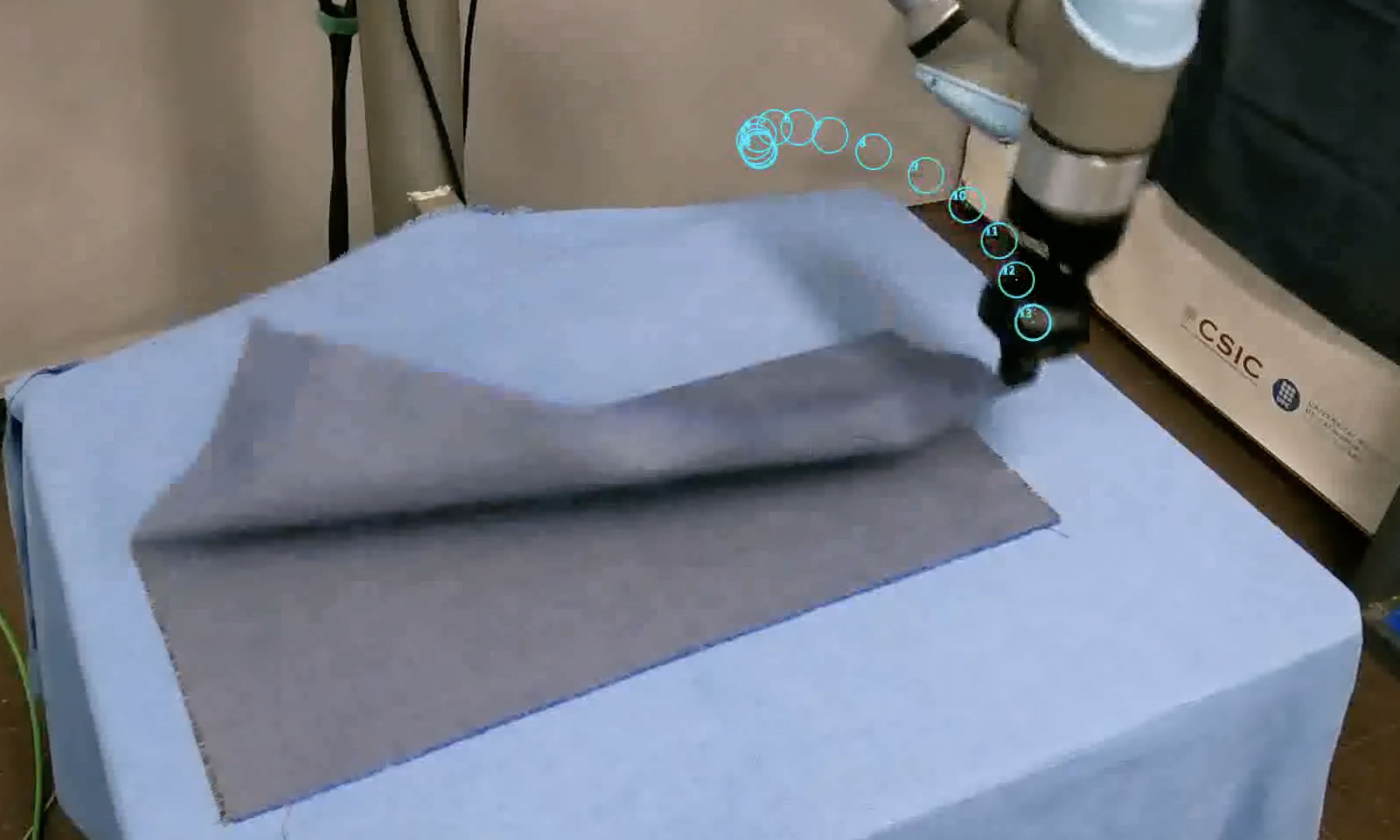}
\end{subfigure}\hfill
\begin{subfigure}{.249\linewidth}
\centering
\includegraphics[width=\linewidth]{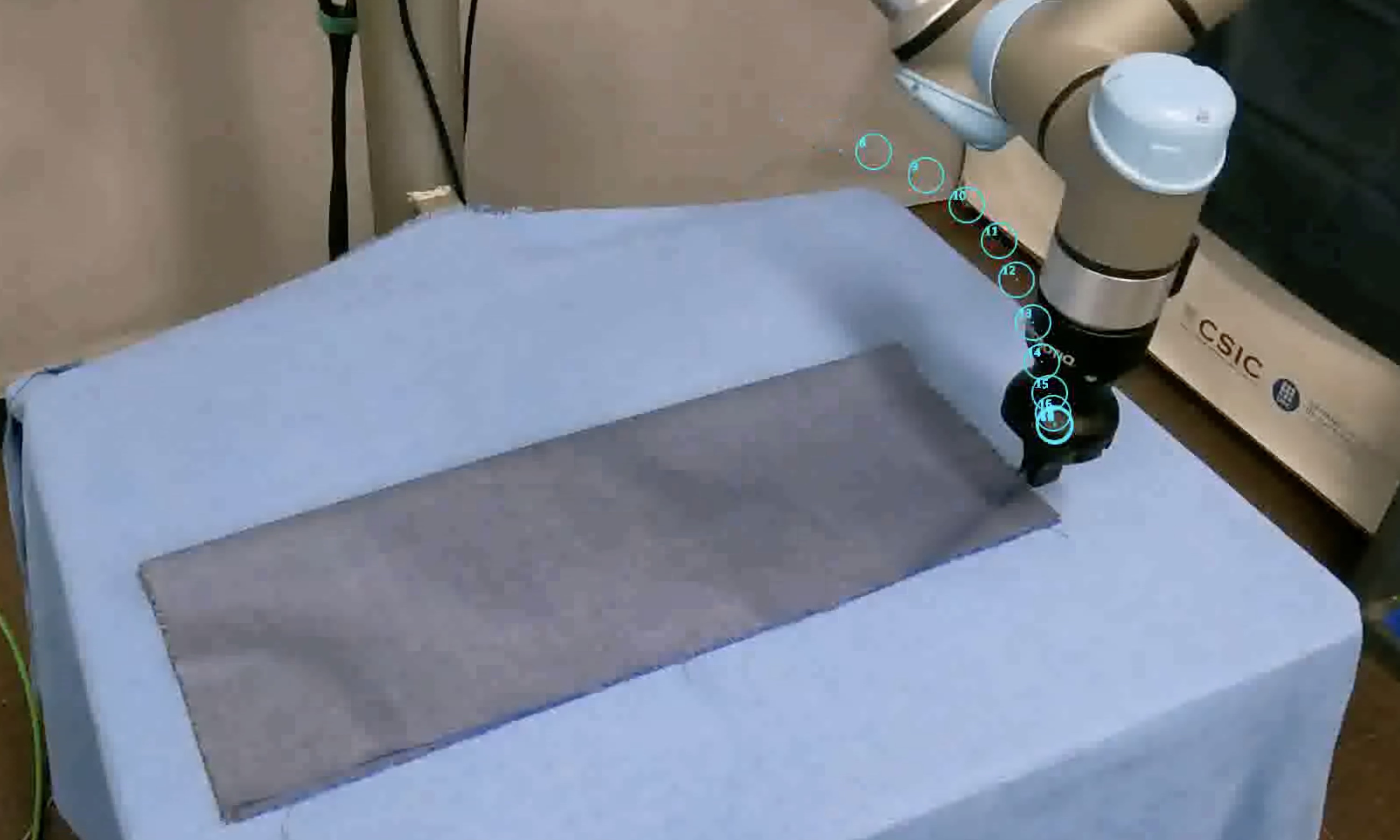}
\end{subfigure}
\caption{An illustrative example of a folding trajectory generated by our MPC pipeline, with a woolen piece of cloth. The trajectory performed by the gripper is displayed as numbered cyan circles.}
\label{fig:illustrative_fold}
\end{figure*}

\subsection{Simulation Experiments}
As a start, we consider an assessment of the proposed LMPC strategy in simulation. The cloth is modeled as a rectangle with size 59 cm $\times$ 42 cm, discretized as a mesh of size $17 \times 13$ nodes. 
The initial flat pose of the cloth is obtained through a motion capture system, as detailed in \cite{coltraro2025tracking}. As discussed in Section \ref{sec:controller_design}, our robot manipulates the cloth by keeping a fixed orientation of the end-effector, which is orthogonal to the table the cloth is placed on. We compute the dynamics in \eqref{eq:koopman_dynamics} by using 100 training trajectories of length 1.5 s, with a sampling time of $\Delta t =0.01 $s. The training folds happen in the form of parabolic trajectories, tilted towards the center of the cloth. \new{While the use of parabolic trajectories introduces a bias in the training data, there is consensus that such trajectory shapes may be effective enough for dynamic folding \cite{li2015folding,hietala2022learning}. Moreover, as we discussed in Section \ref{sec:controller_design}, the synthesized MPC trajectories are not constrained to have the shape of parabolas.} 

In order to simulate the fixed orientation of the end-effector, we simulate the grasping by controlling the corner and the node on the boundary directly adjacent to it. A visualization of some training trajectories is offered in Fig.\ \ref{fig:training_parabolas}.
The simulation pipeline is implemented in \texttt{Matlab}, and the OCP is solved online by means of the \texttt{YALMIP} library.

\new{Note that, in order to improve the generalization capability of our pipeline, the training set may be augmented by additional data in which, e.g., the initial pose of the cloth is perturbed. On the other hand, if different cloth materials or shapes are employed, the SOM's parameters described in Section \ref{sec:case_study} would need to be updated accordingly, our approach being model-based.}
\subsubsection{Impact of the size of the surrogate state} We firstly aim at assessing how the folding error, computed \emph{at the end} of a simulation, varies with $m$, i.e., the size of the surrogate state $\zvec$. In this experiment, the MPC prediction horizon is set to 50 time-steps. The folding error is obtained by calculating the Euclidean distance between the meshes, after aligning the target mesh and the one obtained through MPC with a minimal SVD rotation. As shown in Fig.\ \ref{fig:state_size_effect}, quality of the folds improves for larger values of $m$. 
\subsubsection{Impact of the prediction horizon} Choosing a large enough size of the surrogate state $\zvec$, we are now interested on how the size of the MPC prediction window affects the trajectory generation. As illustrated in Fig.\ \ref{subfig:control_trajs_pred_hor}, when the horizon is short, the trajectories do not go towards the center of the cloth in the beginning. Conversely, this behavior emerges for larger values of the MPC prediction horizon. We empirically observed that this is a relevant feature for the fold to be successful, as confirmed by Fig.\ \ref{subfig:boxplots_mesh_error_pred_hor}.
\subsubsection{Running cost} Next, we also consider the performance of the generated trajectories in terms of the following cumulative cost index, computed on the simulation horizon $T_\textrm{max}$, for $Q' = 1\ \textrm{m}^{-2}$, $R=500\ \textrm{m}^{-2}$:
\begin{equation}
    \mathcal J = \sum_{\kappa=0}^{T_\textrm{max}}(\xvec_\kappa - \xvec_r)^\intercal\mathbf Q'(\xvec_\kappa-\xvec_r) + \Delta \uvec_\kappa^\intercal \mathbf R \Delta\uvec_\kappa.\label{eq:performance_index}
\end{equation}
 This cost can be viewed as the cost our model predictive controller tries to minimize, in a greedy fashion (by means of the receding horizon approach), based on the forecasts of the surrogate model, i.e., $\xvec_t \approx \mathfrak C\zvec_t$, and the constraints on the control action described in Section \ref{sec:controller_design}. Fig.\ \ref{subfig:boxplots_running_cost} shows that the cost index $\mathcal J$ grows at a sublinear rate compared to the size of the simulation horizon $T_{\textrm{max}}$\footnote{Note that, in our implementation, the MPC is started at $T_{\textrm{max}}=30$, in order for the SOM to reach a stable initial state.}.
 \begin{figure*}[t]
\centering
\includegraphics[width=.19\linewidth]{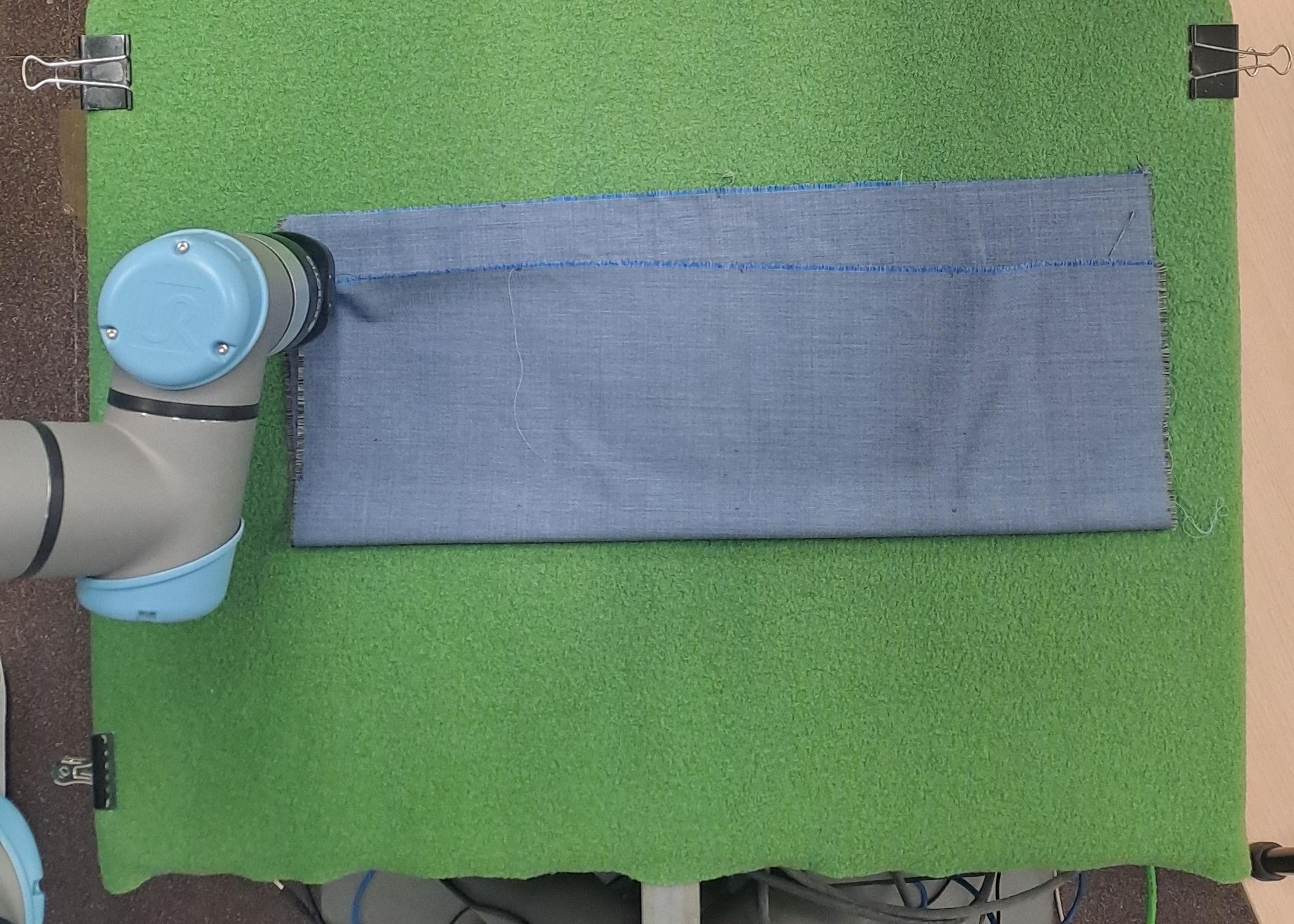}
\includegraphics[width=.19\linewidth]{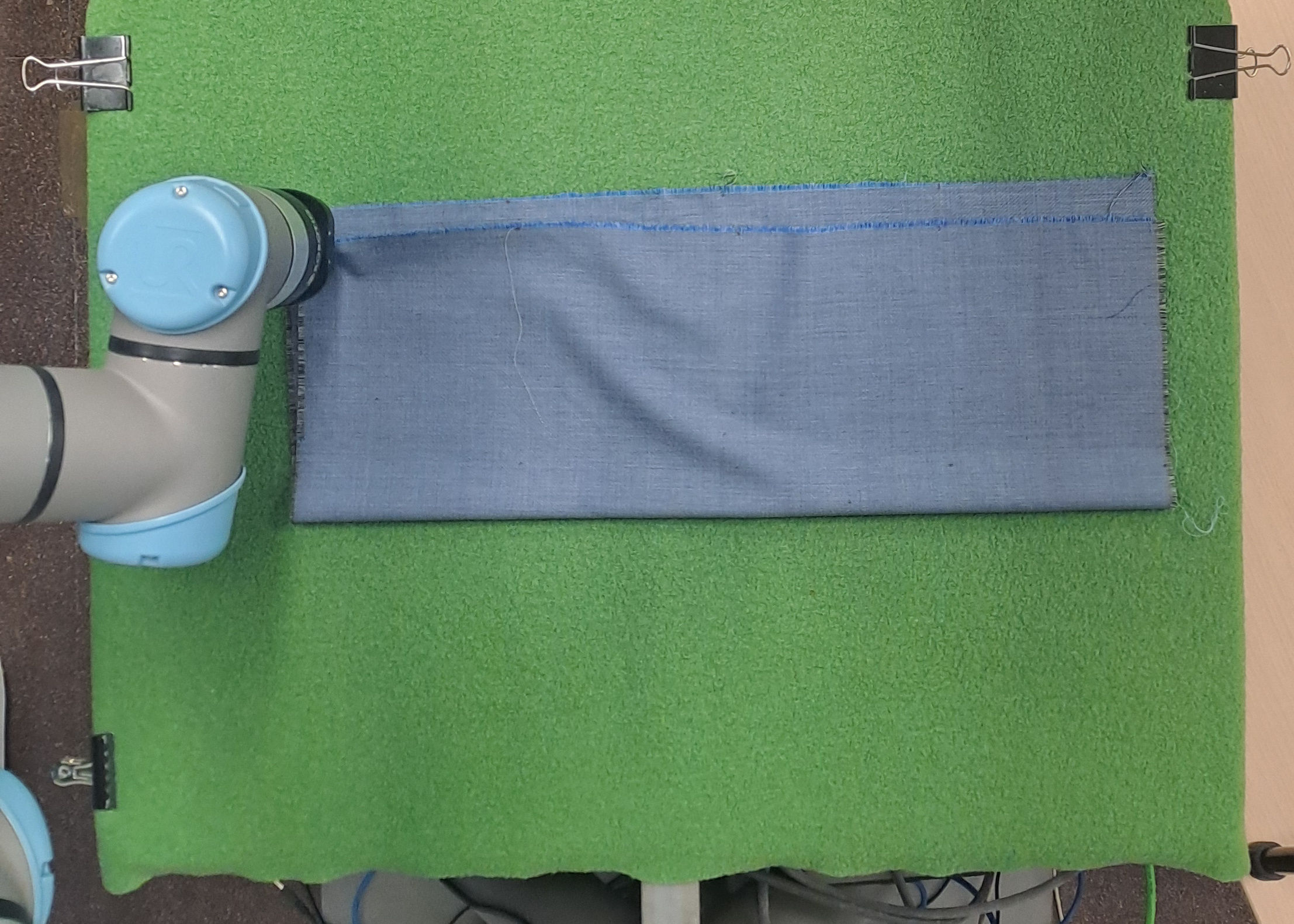}
\includegraphics[width=.19\linewidth]{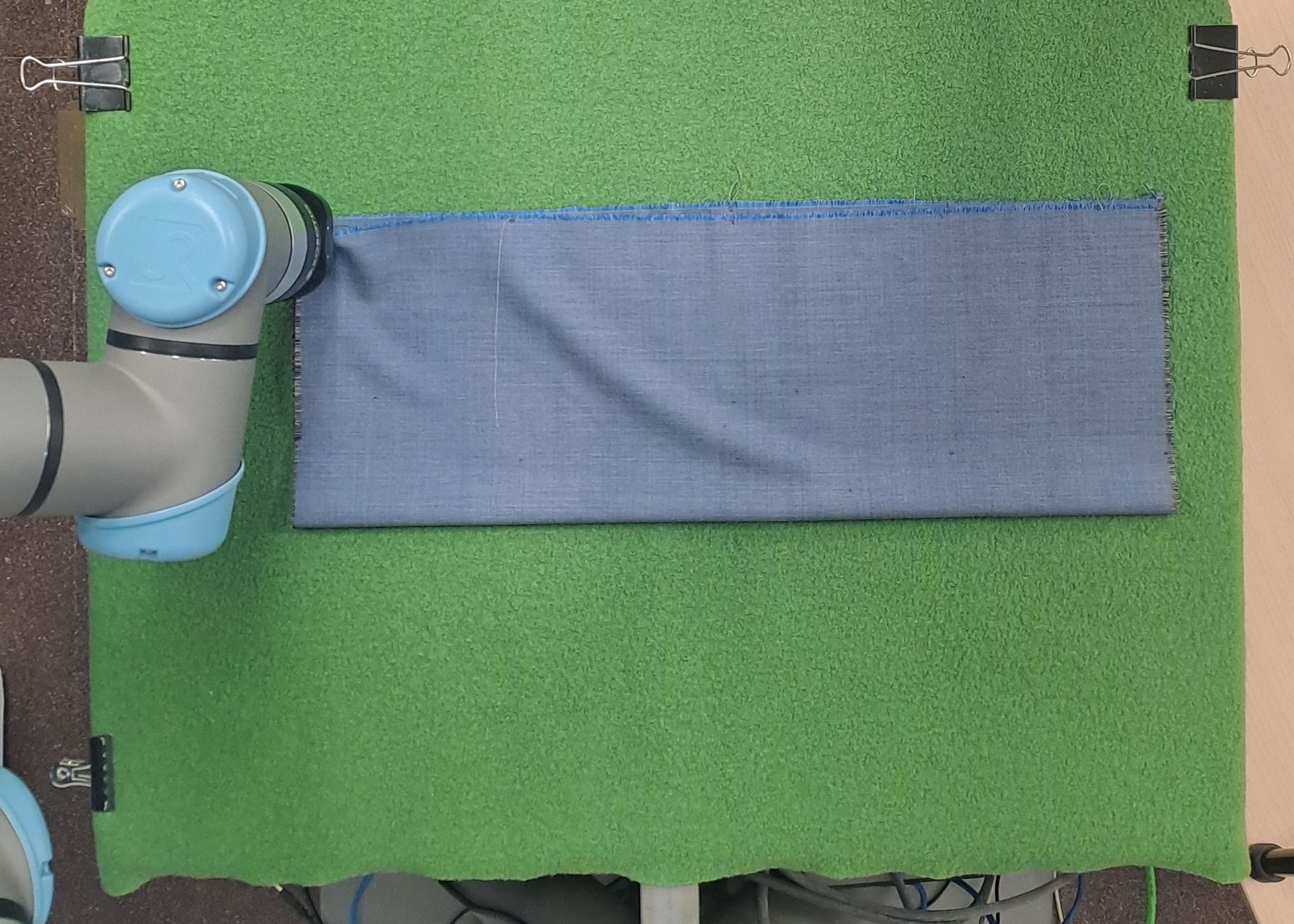}
\includegraphics[width=.19\linewidth]{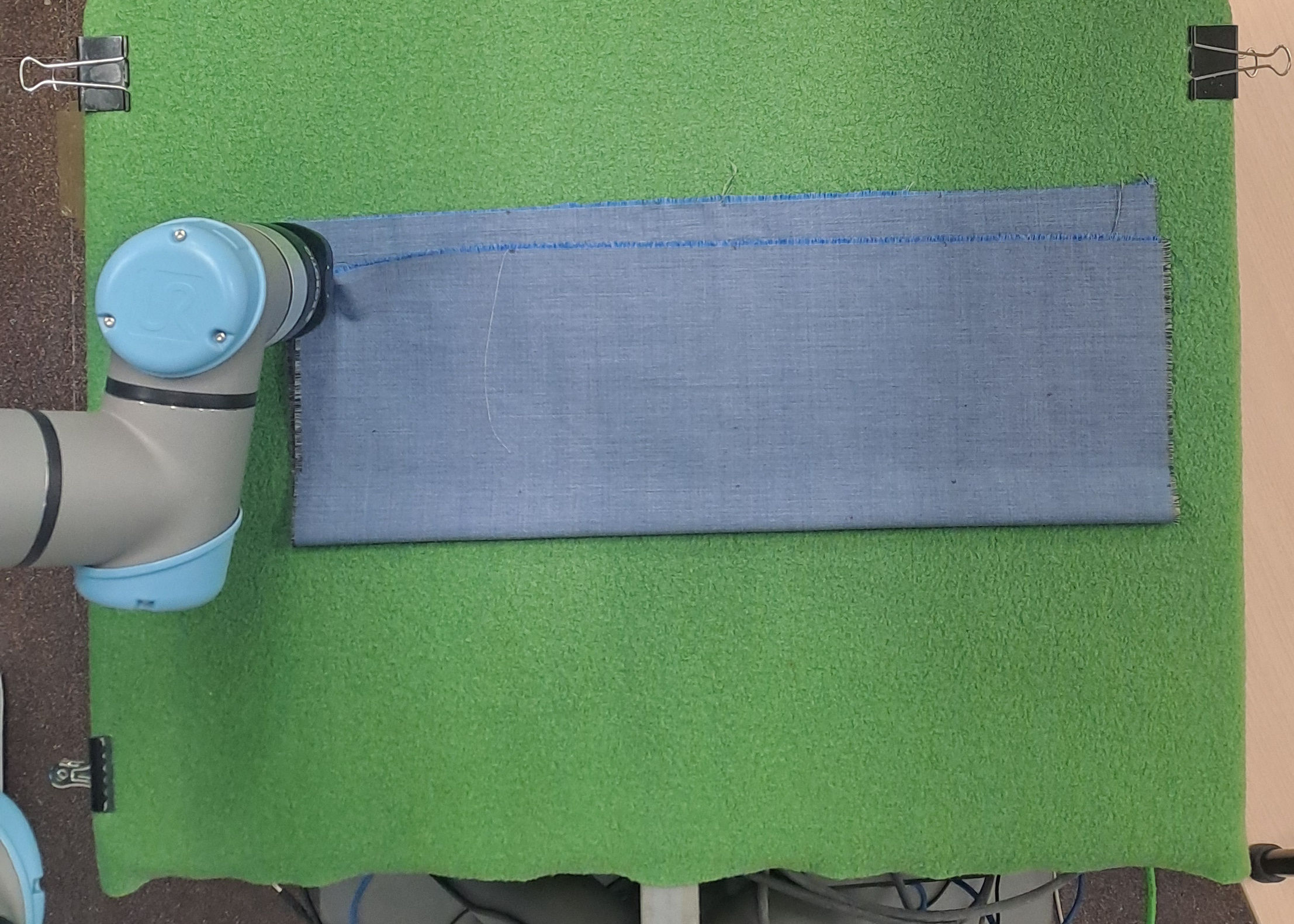}
\includegraphics[width=.19\linewidth]{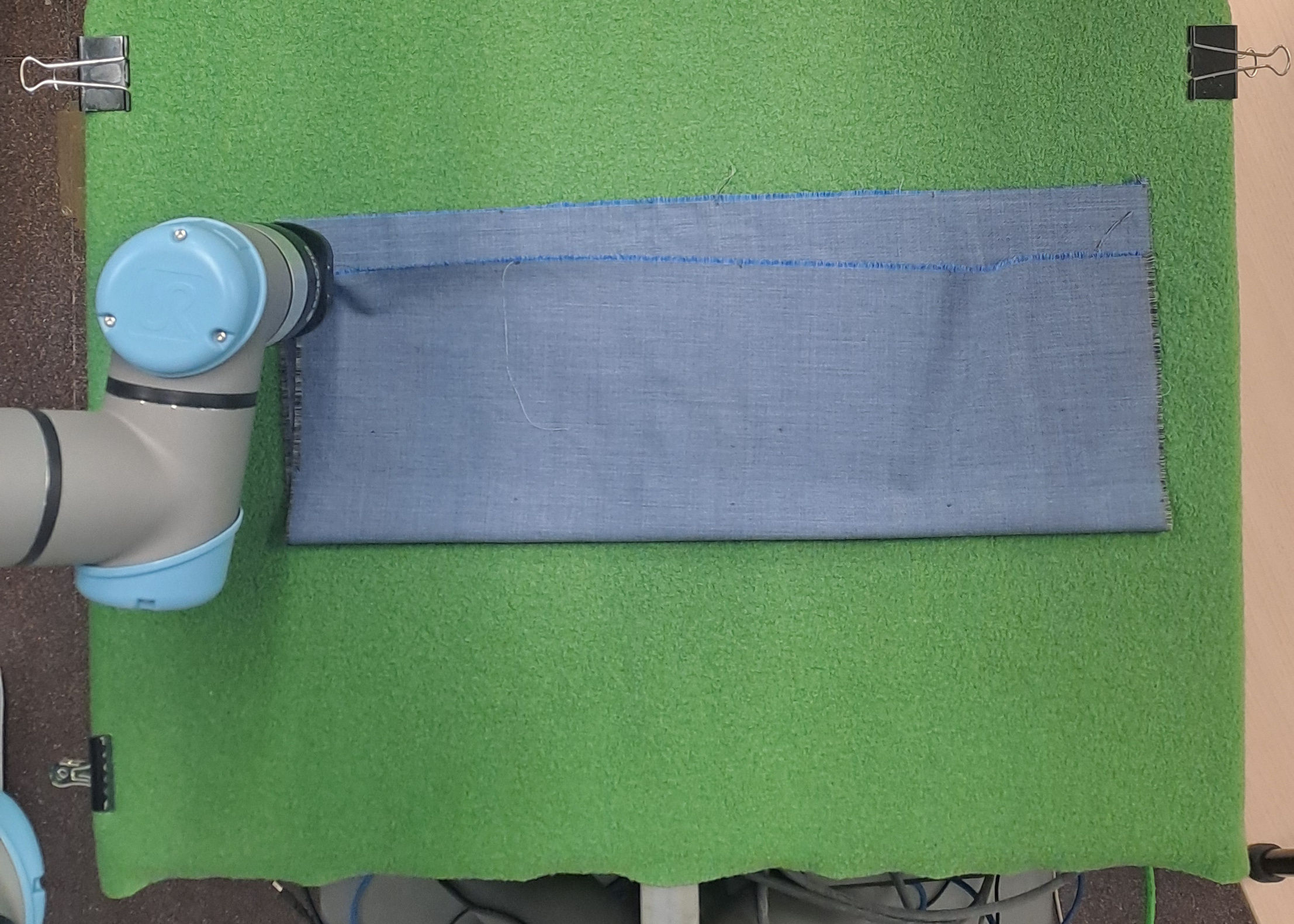}\\
\vspace{.1cm}
\includegraphics[width=.19\linewidth]{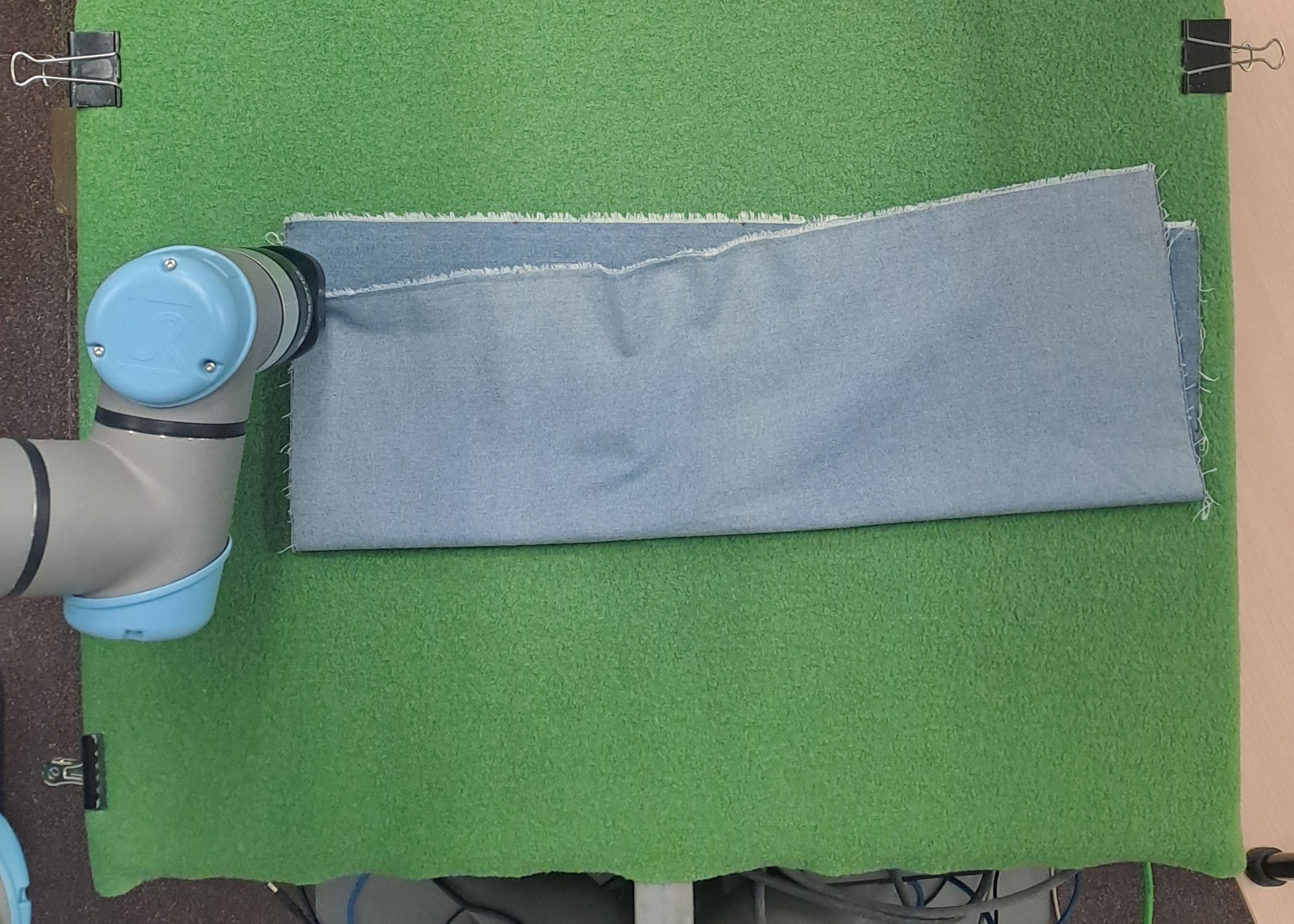}
\includegraphics[width=.19\linewidth]{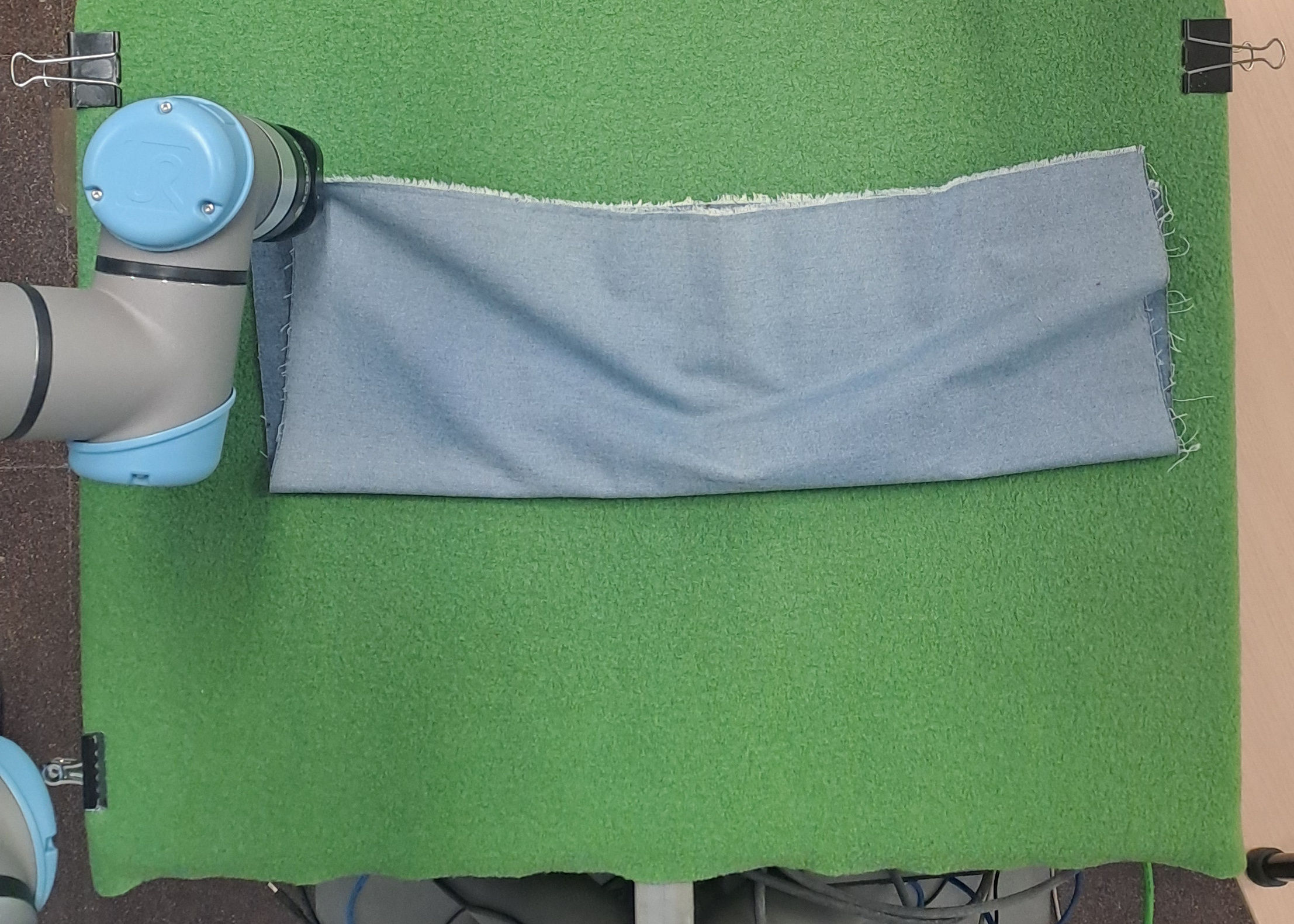}
\includegraphics[width=.19\linewidth]{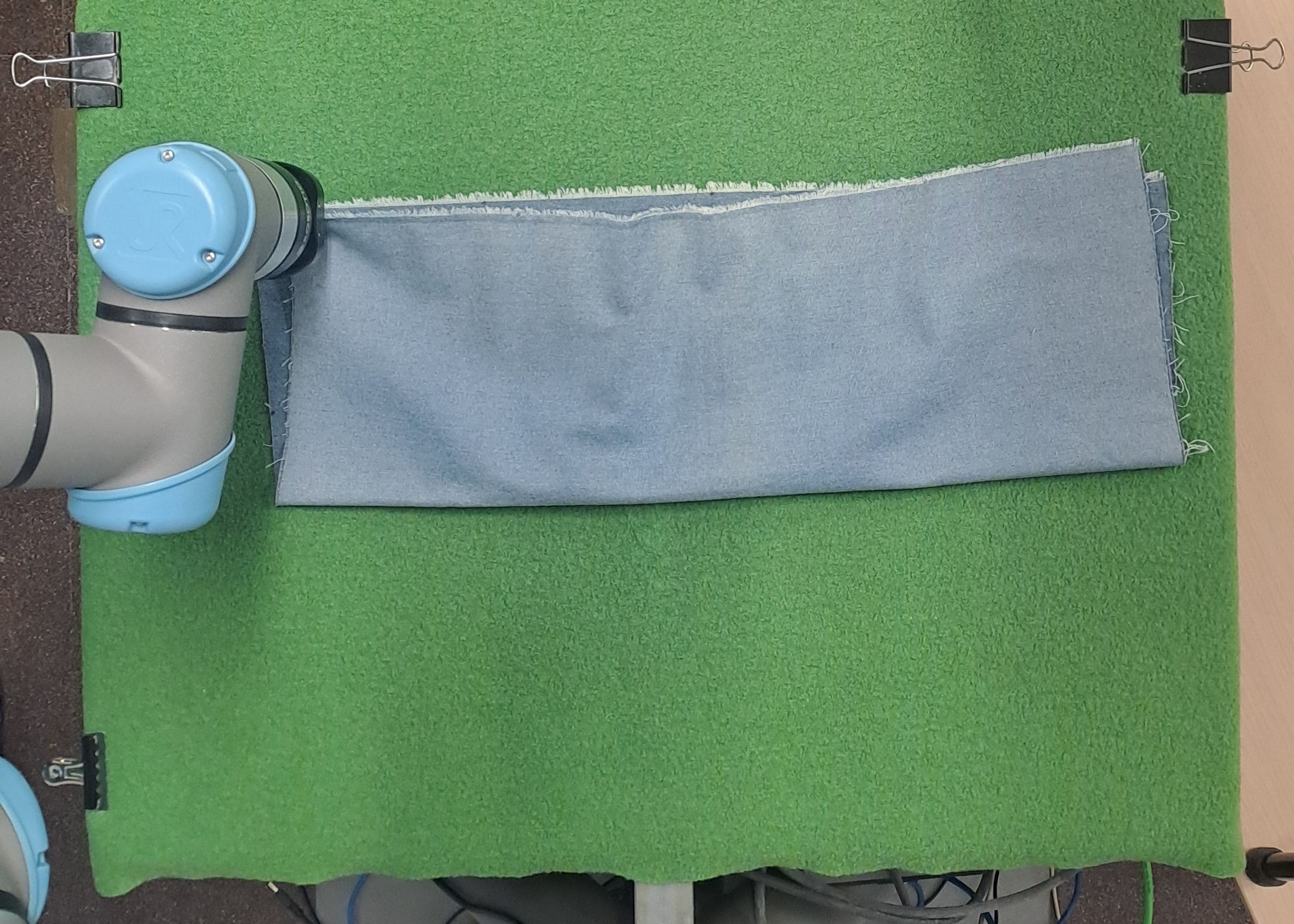}
\includegraphics[width=.19\linewidth]{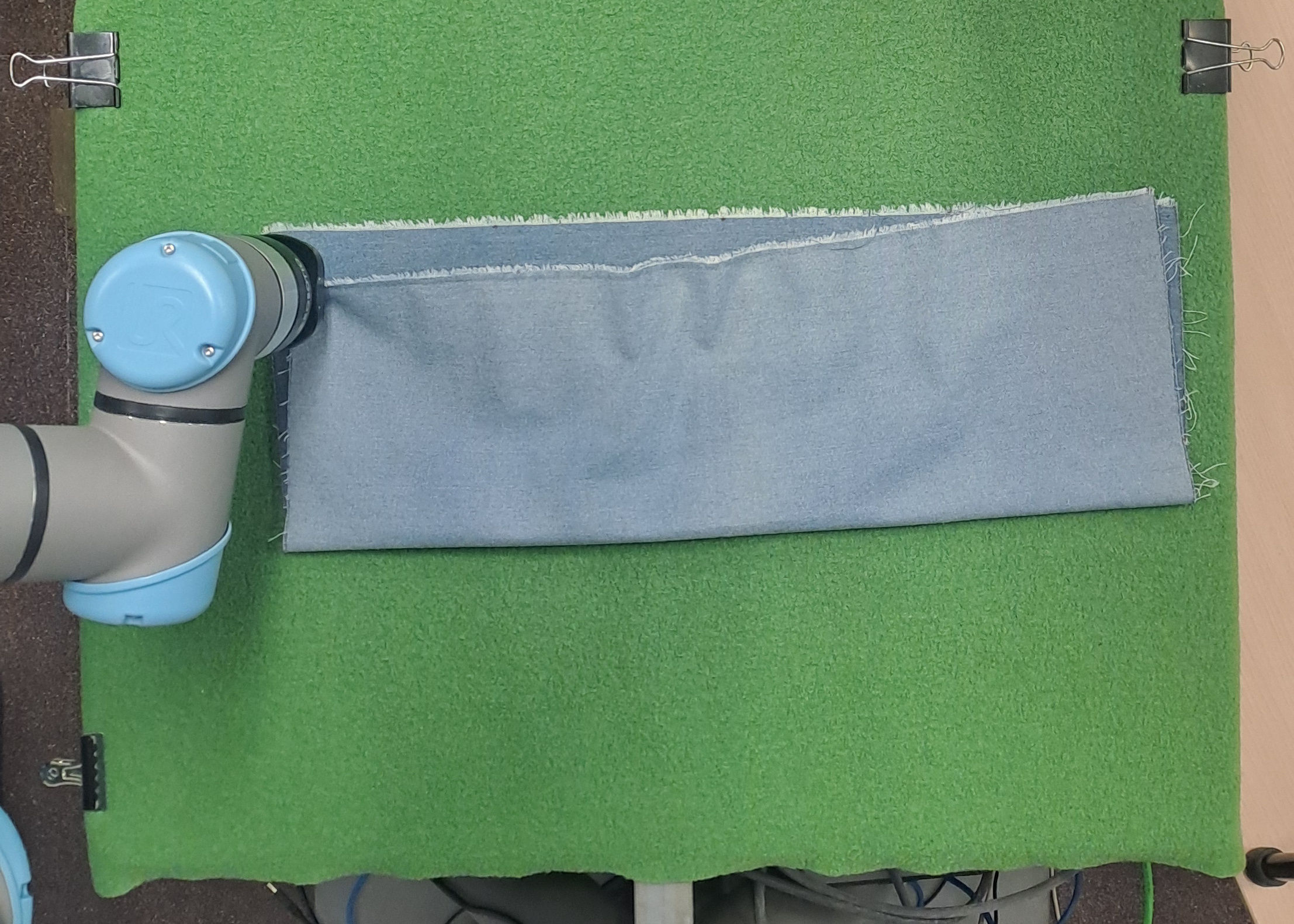}
\includegraphics[width=.19\linewidth]{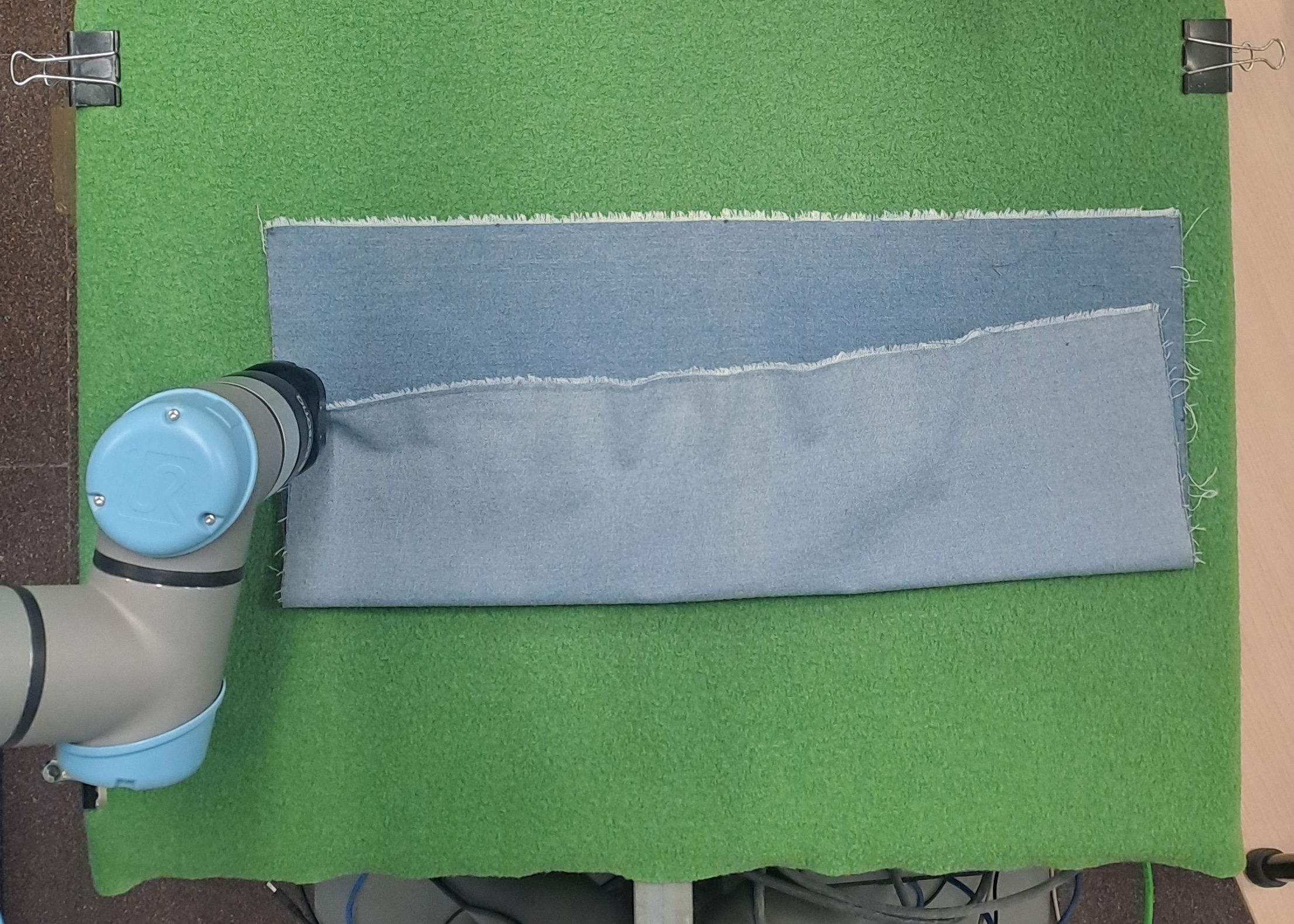}\\
\includegraphics[angle=90, width=.19\linewidth]{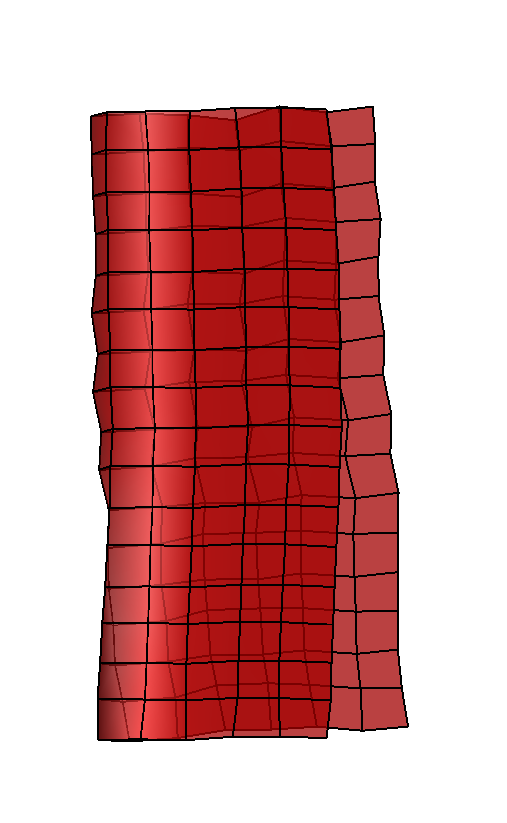}
\includegraphics[angle=90, width=.19\linewidth]{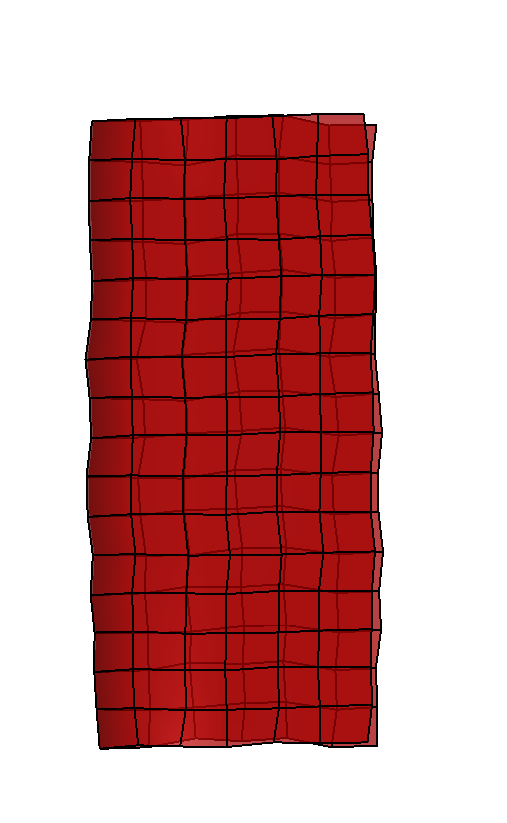}
\includegraphics[angle=90, width=.19\linewidth]{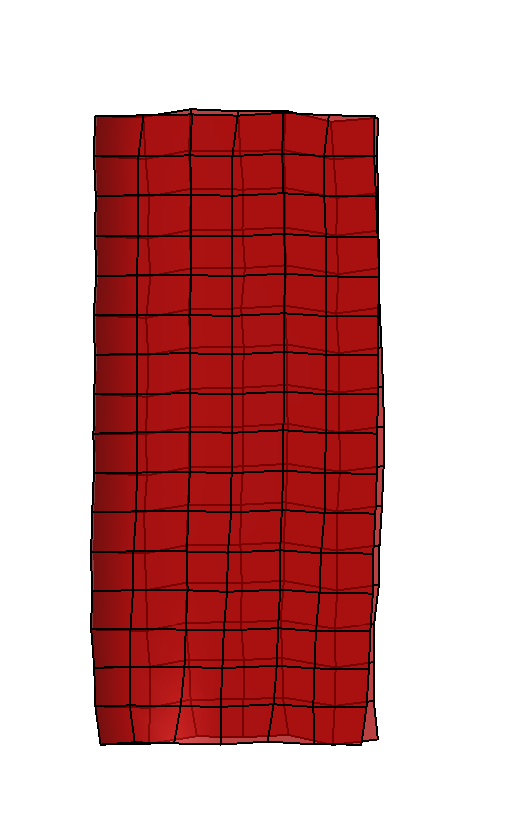}
\includegraphics[angle=90, width=.19\linewidth]{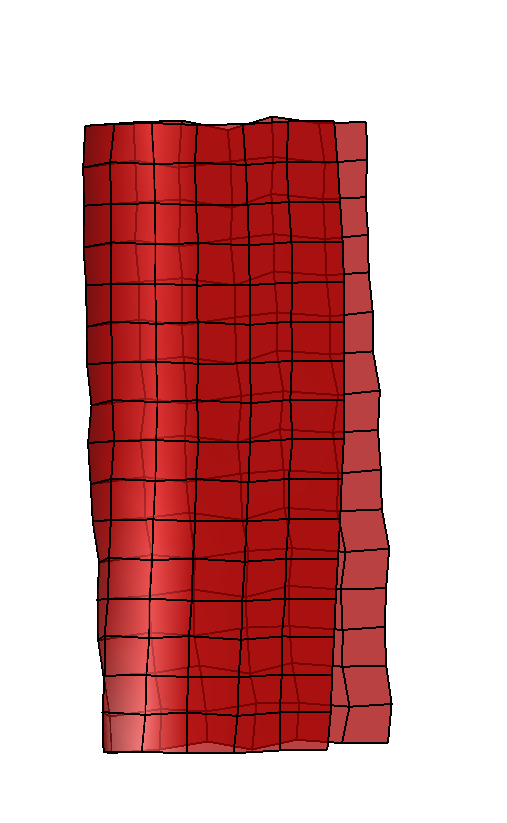}
\includegraphics[angle=90, width=.19\linewidth]{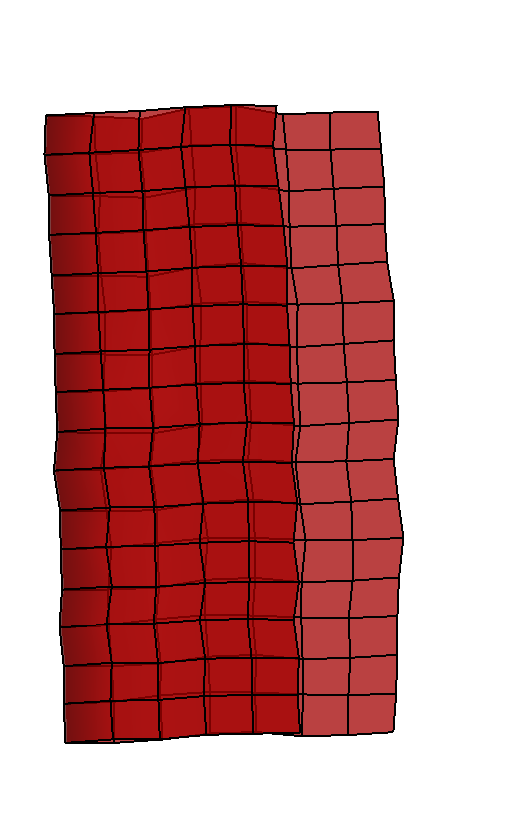}
\caption{A qualitative visualization of the folds obtained with our MPC pipeline, in the sim-to-real evaluation. The top row contains the results for wool, the second row the results for denim, and the bottom row the target folds. }
\label{fig:real_fold}
\end{figure*}
\subsection{Sim-to-Real Evaluation}
\begin{table}
    \centering
    \caption{Parameters of the cloths employed in our folding setup.}
    \begin{tabularx}{\linewidth}{Xcccc}
    \toprule
    & \tableheadline{Size [cm\textsuperscript{2}]}& \tableheadline{Material} & \tableheadline{Density [kg/m\textsuperscript{2}]} & \tableheadline{Example}\\
    \midrule 
         {Cloth I:} & $59\times42$ &Wool &0.1804 & Suit\\
         {Cloth II:}&$59\times42$& Denim &0.3046 & Jeans\\
         \bottomrule
    \end{tabularx}
    \label{table:sota}
\end{table}
We can now consider how the trajectories generated with our pipeline transfer to a real robot setup. The goal is again to fold a rectangular piece of cloth by controlling only one of its corners. {We consider two different garments, whose parameters can be found in Table \ref{table:sota}. \new{The chosen materials are representative of common garments that can be encountered in household scenarios, and they differ both in terms of stiffness and density, allowing to test the sim-to-real performance of our algorithm in diverse scenarios.}}
\new{Given the high speed of the motions of interest (with a smaller duration than 1.5 s), the trajectories are executed in open loop, as the processing of high-frequency real-time feedback would give a delay to the controller, i.e.: current state-of-the-art visual feedback techniques for cloth perception would provide data at a low frequency with a high delay, compared to the speed of the motion.} Closing the loop would also require additional compensation and simplifications of the model, see, e.g., \cite{luque2024model,caldarelli2023quadratic}. {The trajectories are executed with a controller in Cartesian space.}
In this experiment, we train the two data-driven models on 100 trajectories of duration 1.5 s, generated with single-handed motions. The testing folded poses are generated by running 5 bi-manual, quasi-static manipulation trajectories with the SOM. Our experimentation platform is a \texttt{UR5} robotic arm that grasps the cloth with a fixed, vertical gripper orientation, as discussed in Section \ref{sec:controller_design}. Fig.\ \ref{fig:illustrative_fold} shows an illustrative folding trajectory generated by our MPC pipeline\footnote{Gripper tracking realized with the \texttt{Tracker} software.}. 

\subsubsection{Performance} {In order to assess the performance of the generated trajectories, we compute the \emph{folding ratio}, i.e., the ratio between the area of visible surface of the cloth after folding with the real robot, and the area of the unfolded surface. This value, denoted as $\mathcal R_\textrm{real}$, is compared against the ratio between the area of the target visible surface from the simulator, and the unfolded surface from the simulator. The latter ratio is denoted as $\mathcal R_\textrm{target}$. The trajectories with the real robot are repeated 2 times each. Table \ref{table:sim2real} shows the associated key performance indicator (KPI) $\mathcal E_\textrm{fold}$, computed as\looseness=-1
\begin{equation}
    \mathcal E_\textrm{fold}= \frac{\lvert\mathcal R_\textrm{real} - \mathcal R_\textrm{target}\rvert}{\mathcal R_\textrm{target}}.\label{eq:sim2real_kpi}
\end{equation}
\new{Since this KPI may not be able to capture nuanced features of the achieved fold (e.g., in terms of exact placement of the cloth's corner), we supply a zenithal visualization of such folds in Fig.\ \ref{fig:real_fold}. Indeed, while the denim folds achieve a lower KPI than the woolen ones, we can observe in Fig.\ \ref{fig:real_fold} that the surface of the wool garment is more flat as opposed to the denim one, and the corner placement is in turn more accurate w.r.t.\ the reference fold. This is due to the different physical properties of the garments.} 

\new{Overall}, our algorithm achieves a satisfactory sim-to-real performance, being able to fold both garments to a given target pose.}

\begin{table}
    \centering
    \caption{Relative folding error $\mathcal E_\textrm{fold}$ between the real and target cloth folds according to \eqref{eq:sim2real_kpi}, for 5 target folds. The values are averaged across 2 repetitions of each trajectory.}
    \label{table:sim2real}
    \setlength{\tabcolsep}{9pt}
\renewcommand{\arraystretch}{1}
    \begin{tabularx}{\linewidth}{Xccccc}
    \toprule
    \tableheadline{Material} & & & & & \\
    \midrule
    Wool &6.02\% &3.99\% &2.21\% &6.44\% & 7.23\%\\
    Denim &3.97\%&1.76\% &0.769\% &4.34\% &3.81\%\\
    
         \bottomrule
    \end{tabularx}
    \label{tab:my_label}
    \label{table:perturbed_init_cond}
\end{table}

\section{Conclusions}
\label{sec:conclusions}

{In this paper, we have shown how accurate, physics-based modeling of cloth can be used to generate a linear, surrogate model by means of Koopman operator regression. 
This surrogate model, coupled with suitable constraints, is then embedded in a model predictive control pipeline in closed loop with the physics-aware one, to generate folding trajectories to be executed on a real robot.}

\new{Under the assumption that the cloth is initially flat on a table and that we have performed system identification with a physical simulator, the Koopman model has been trained \emph{exclusively on simulated data}, which has been shown here to be realistic enough to achieve a successful zero-shot simulation-to-reality transfer of the generated high-speed trajectories.}{The non-linear physical model is accurate as it accounts for \emph{aerodynamics} and \emph{inextensibility}. Moreover, the Koopman-based linearization of the cloth's dynamics has a twofold benefit. On one hand, it allows to compute the folding trajectories in a \emph{derivative-free} manner, w.r.t.\ the non-linear simulator. Moreover, the dimensionality of the cloth's state space is reduced after the Koopman lifting, w.r.t.\ the original state space. Overall, these points allow to solve efficient linear quadratic optimal control problems.}

{The model predictive control strategy proposed in this paper opens the way to interesting future research, aimed, for instance, at including feedback from the real cloth status. Such feedback could be employed, e.g., to align the robot with the cloth in an initial phase, and perform \emph{grasping} primitives, so as to start the folding motion according to the simulated environment. In terms of grasping, we further showed that a vertical grasp is enough to achieve successful cloth folds, provided that it is modeled correctly in the simulator. Nonetheless, different types of grasps could be studied and developed, to account for different cloth shapes and parameters.}

\bibliographystyle{ieeetr}
\bibliography{refs}

@article{caldarelli2025linear,
  title={Linear quadratic control of nonlinear systems with {K}oopman operator learning and the {N}ystr{\"o}m method},
  author={Caldarelli, Edoardo and Chatalic, Antoine and Colom{\'e}, Adri{\`a} and Molinari, Cesare and Ocampo-Martinez, Carlos and Torras, Carme and Rosasco, Lorenzo},
  journal={Automatica},
  volume={177},
  pages={112302},
  year={2025},
  publisher={Elsevier}
}

@inproceedings{li2015folding,
  title={Folding deformable objects using predictive simulation and trajectory optimization},
  author={Li, Yinxiao and Yue, Yonghao and Xu, Danfei and Grinspun, Eitan and Allen, Peter K},
  booktitle={2015 IEEE/RSJ International Conference on Intelligent Robots and Systems (IROS)},
  pages={6000--6006},
  year={2015},
  organization={IEEE}
}

@article{bevanda2021koopman,
  title={Koopman operator dynamical models: Learning, analysis and control},
  author={Bevanda, Petar and Sosnowski, Stefan and Hirche, Sandra},
  journal={Annual Reviews in Control},
  volume={52},
  pages={197--212},
  year={2021},
  publisher={Elsevier}
}

@article{coltraro2025aero,
  title = {A practical aerodynamic model for dynamic textile manipulation in robotics},
  journal = {Mechanism and Machine Theory},
  volume = {209},
  pages = {105993},
  year = {2025},
  issn = {0094-114X},
  doi = {https://doi.org/10.1016/j.mechmachtheory.2025.105993},
  url = {https://www.sciencedirect.com/science/article/pii/S0094114X25000825},
  author = {Coltraro, Franco and Amorós, Jaume and Torras, Carme and Alberich-Carramiñana, Maria},
}

@article{coltraro2025tracking,
  title={Tracking cloth deformation: A novel dataset for closing the sim-to-real gap for robotic cloth manipulation learning},
  author={Coltraro, Franco and Borr{\`a}s, J{\'u}lia and Alberich-Carrami{\~n}ana, Maria and Torras, Carme},
  journal={The International Journal of Robotics Research},
  pages={02783649251317617},
  year={2025},
  publisher={SAGE Publications Sage UK: London, England}
}

@article{coltraro2022inextensible,
  title={An inextensible model for the robotic manipulation of textiles},
  author={Coltraro, Franco and Amor{\'o}s, Jaume and Alberich-Carrami{\~n}ana, Maria and Torras, Carme},
  journal={Applied Mathematical Modelling},
  volume={101},
  pages={832--858},
  year={2022},
  publisher={Elsevier}
}

@inproceedings{Bergou:2006:QBM,
 author = {Bergou, Miklos and Wardetzky, Max and Harmon, David and Zorin, Denis and Grinspun, Eitan},
 title = {A Quadratic Bending Model for Inextensible Surfaces},
 booktitle = {Proceedings of the Fourth Eurographics Symposium on Geometry Processing},
 series = {SGP '06},
 year = {2006},
 isbn = {3-905673-36-3},
 location = {Cagliari, Sardinia, Italy},
 pages = {227--230},
 numpages = {4},
 acmid = {1281987},
 publisher = {Eurographics Association},
}

@book{Zienkiewicz:2005:FEM,
  author = {Zienkiewicz, Olgierd C. and Taylor, Robert L. and Zhu, Jizhong Z.},
  isbn = {0750663200},
  month = may,
  publisher = {Butterworth-Heinemann},
  shorttitle = {The Finite Element Method},
  title = {The Finite Element Method: Its Basis and Fundamentals, Sixth Edition},
  year = 2005
}

@article{yin2021modeling,
  title={Modeling, learning, perception, and control methods for deformable object manipulation},
  author={Yin, Hang and Varava, Anastasia and Kragic, Danica},
  journal={Science Robotics},
  volume={6},
  number={54},
  pages={eabd8803},
  year={2021},
  publisher={American Association for the Advancement of Science}
}

@article{longhini2024unfolding,
  title={Unfolding the literature: A review of robotic cloth manipulation},
  author={Longhini, Alberta and Wang, Yufei and Garcia-Camacho, Irene and Blanco-Mulero, David and Moletta, Marco and Welle, Michael and Aleny{\`a}, Guillem and Yin, Hang and Erickson, Zackory and Held, David and others},
  journal={Annual Review of Control, Robotics, and Autonomous Systems},
  volume={8},
  year={2024},
  publisher={Annual Reviews}
}

@inproceedings{ha2022flingbot,
  title={Flingbot: The unreasonable effectiveness of dynamic manipulation for cloth unfolding},
  author={Ha, Huy and Song, Shuran},
  booktitle={Conference on Robot Learning},
  pages={24--33},
  year={2022},
  organization={PMLR}
}

@inproceedings{hietala2022learning,
  title={Learning visual feedback control for dynamic cloth folding},
  author={Hietala, Julius and Blanco--Mulero, David and Alcan, Gokhan and Kyrki, Ville},
  booktitle={2022 IEEE/RSJ International Conference on Intelligent Robots and Systems (IROS)},
  pages={1455--1462},
  year={2022},
  organization={IEEE}
}

@article{longhini2024adafold,
  title={Adafold: Adapting folding trajectories of cloths via feedback-loop manipulation},
  author={Longhini, Alberta and Welle, Michael C and Erickson, Zackory and Kragic, Danica},
  journal={IEEE Robotics and Automation Letters},
  year={2024},
  publisher={IEEE}
}

@article{blanco2024benchmarking,
  title={Benchmarking the sim-to-real gap in cloth manipulation},
  author={Blanco-Mulero, David and Barbany, Oriol and Alcan, Gokhan and Colom{\'e}, Adri{\`a} and Torras, Carme and Kyrki, Ville},
  journal={IEEE Robotics and Automation Letters},
  volume={9},
  number={3},
  pages={2981--2988},
  year={2024},
  publisher={IEEE}
}

@inproceedings{caldarelli2023quadratic,
  title={Quadratic Dynamic Matrix Control for Fast Cloth Manipulation},
  author={Caldarelli, Edoardo and Colom{\'e}, Adri{\`a} and Ocampo-Martinez, Carlos and Torras, Carme},
  booktitle={2023 IEEE/RSJ International Conference on Intelligent Robots and Systems (IROS)},
  pages={8178--8185},
  year={2023},
  organization={IEEE}
}

@article{luque2024model,
  title={Model predictive control for dynamic cloth manipulation: Parameter learning and experimental validation},
  author={Luque, Adri{\`a} and Parent, David and Colom{\'e}, Adri{\`a} and Ocampo-Martinez, Carlos and Torras, Carme},
  journal={IEEE Transactions on Control Systems Technology},
  year={2024},
  publisher={IEEE}
}

@inproceedings{blanco2023qdp,
  title={{QDP}: Learning to sequentially optimise quasi-static and dynamic manipulation primitives for robotic cloth manipulation},
  author={Blanco-Mulero, David and Alcan, Gokhan and Abu-Dakka, Fares J and Kyrki, Ville},
  booktitle={2023 IEEE/RSJ International Conference on Intelligent Robots and Systems (IROS)},
  pages={984--991},
  year={2023},
  organization={IEEE}
}

@article{nystrom1930praktische,
  title={{\"U}ber die praktische {A}ufl{\"o}sung von {I}ntegralgleichungen mit {A}nwendungen auf {R}andwertaufgaben},
  author={{N}ystr{\"o}m, Evert J},
  year={1930}
}

@article{williams2000using,
  title={Using the {N}ystr{\"o}m method to speed up kernel machines},
  author={Williams, Christopher and Seeger, Matthias},
  journal={Advances in neural information processing systems},
  volume={13},
  year={2000}
}

@article{koopman1931hamiltonian,
  title={Hamiltonian systems and transformation in {H}ilbert space},
  author={{K}oopman, Bernard O},
  journal={Proceedings of the National Academy of Sciences},
  volume={17},
  number={5},
  pages={315--318},
  year={1931},
  publisher={National Acad Sciences}
}

@inproceedings{avigal2022speedfolding,
  title={Speedfolding: Learning efficient bimanual folding of garments},
  author={Avigal, Yahav and Berscheid, Lars and Asfour, Tamim and Kr{\"o}ger, Torsten and Goldberg, Ken},
  booktitle={2022 IEEE/RSJ International Conference on Intelligent Robots and Systems (IROS)},
  pages={1--8},
  year={2022},
  organization={IEEE}
}

@inproceedings{mason1993dynamic,
  title={Dynamic manipulation},
  author={Mason, Matthew T and Lynch, Kevin M},
  booktitle={Proceedings of 1993 IEEE/RSJ International Conference on Intelligent Robots and Systems (IROS'93)},
  volume={1},
  pages={152--159},
  year={1993},
  organization={IEEE}
}

@article{coltraro2024novel,
  title={A novel collision model for inextensible textiles and its experimental validation},
  author={Coltraro, Franco and Amor{\'o}s, Jaume and Alberich-Carrami{\~n}ana, Maria and Torras, Carme},
  journal={Applied Mathematical Modelling},
  volume={128},
  pages={287--308},
  year={2024},
  publisher={Elsevier}
}

@article{korda2018linear,
  title={Linear predictors for nonlinear dynamical systems: {K}oopman operator meets model predictive control},
  author={Korda, Milan and Mezi{\'c}, Igor},
  journal={Automatica},
  volume={93},
  pages={149--160},
  year={2018},
  publisher={Elsevier}
}
\end{document}